\documentclass[a4paper,fleqn]{cas-sc}

\usepackage[numbers]{natbib}

\def\tsc#1{\csdef{#1}{\textsc{\lowercase{#1}}\xspace}}
\tsc{WGM}
\tsc{QE}

\usepackage[table]{xcolor}
\definecolor{hl}{RGB}{46,139,87}

\begin{document}
\let\WriteBookmarks\relax
\def\floatpagepagefraction{1}
\def\textpagefraction{.001}

\shorttitle{Medical I2I Translation Using Generative Frameworks}    

\shortauthors{Romoli et al.}  

\title [mode = title]{Cross Modality Image Translation In
Medical Imaging Using Generative
Frameworks}  



%

\author[1]{Giulia Romoli}[orcid=0000-0003-1624-1709]
\cormark[1] 
\ead{giulia.romoli@umu.se}
\credit{Conceptualization, Investigation, Methodology, Software, Writing - First draft, Writing – review and editing}
\author[1,2]{Filippo Ruffini}
\credit{Conceptualization, Investigation, Methodology, Software, Writing – review and editing}
\author[1]{Francesco {Di Feola}}
\credit{Conceptualization, Investigation, Methodology, Software}
\author[2]{Alessia Capoccia}
\credit{Data curation, Investigation, Methodology, Software, Writing - First draft, Writing – review and editing}
\author[3,4]{Luca Boldrini}
\credit{Validation}
\author[5,6]{Arturo Chiti}
\credit{Validation}
\author[7]{Renato Cuocolo}
\credit{Validation}
\author[8,9]{Tugba {Akinci D'Antonoli}}
\credit{Validation}
\author[10]{Fatemeh Darvizeh}
\credit{Validation}
\author[11,12]{Marcello {Di Pumpo}}
\credit{Validation}
\author[13]{Bradley J. Erickson}
\credit{Validation}
\author[14]{Liu Fang}
\credit{Validation}
\author[10]{Deborah Fazzini}
\credit{Validation}
\author[15]{Paola Feraco}
\credit{Validation}
\author[5]{Fabrizia Gelardi}
\credit{Validation}
\author[16]{Francesco Gossetti}
\credit{Validation}
\author[10]{Ana Isabel {Hernáiz Ferrer}}
\credit{Validation}
\author[17,18]{Michail E. Klontzas}
\credit{Validation}
\author[19]{Seyedmehdi Payabvash}
\credit{Validation}
\author[20]{Katrine Riklund}
\credit{Validation}
\author[20]{Sara N. Strandberg}
\credit{Validation}
\author[2]{Valerio Guarrasi}
\credit{Conceptualization, Methodology, Supervision, Writing – review and editing}
\author[1,2]{Paolo Soda}[orcid=0000-0003-2621-072X]
\cormark[1]
\ead{paolo.soda@umu.se}
\credit{Conceptualization, Funding acquisition, Project administration, Methodology, Supervision, Writing – review and editing}

\affiliation[1]{organization={Department of Diagnostics and Intervention, Radiation Physics,
Biomedical Engineering, Umeå University},
            addressline={Universitetstorget, 4}, 
            city={Umeå},
            postcode={90187}, 
            country={Sweden}}

\affiliation[2]{organization={Unit of Artificial Intelligence and Computer Systems,
Department of Engineering, Università Campus Bio-Medico di Roma},
            addressline={Via Alvaro del Portillo, 21}, 
            city={Roma},
            postcode={00128}, 
            country={Italy}}

\affiliation[3]{organization={Fondazione Policlinico Universitario "A. Gemelli" IRCCS},
            addressline={Largo Agostino Gemelli, 8}, 
            city={Roma},
            postcode={00168}, 
            country={Italy}}

\affiliation[4]{organization={Università Cattolica del Sacro Cuore},
            addressline={Largo Francesco Vito, 1}, 
            city={Roma},
            postcode={00168 }, 
            country={Italy}}

\affiliation[5]{organization={IRCCS Ospedale San Raffaele},
            addressline={via Olgettina, 60}, 
            city={Milano},
            postcode={20132}, 
            country={Italy}}

\affiliation[6]{organization={Vita-Salute San Raffaele University},
            addressline={Via Olgettina, 58}, 
            city={Milano},
            postcode={20132}, 
            country={Italy}}

\affiliation[7]{organization={Department of Medicine, Surgery and Dentistry, University of Salerno},
            addressline={Via Salvador Allende, 43}, 
            city={Baronissi},
            postcode={84081}, 
            country={Italy}}

\affiliation[8]{organization={Division of Diagnostic and Interventional Neuroradiology, Department of Radiology, University Hospital Basel},
            addressline={Spitalstrasse, 21}, 
            city={Basel},
            postcode={4031}, 
            country={Switzerland}}

\affiliation[9]{organization={Department of Pediatric Radiology, University Children’s Hospital Basel},
            city={Basel},
            country={Switzerland}}            

\affiliation[10]{organization={Centro Diagnostico Italiano (CDI)},
            addressline={Via Simone Saint Bon, 20}, 
            city={Milano},
            postcode={20147}, 
            country={Italy}}

\affiliation[11]{organization={Department of Life Science and Public Health, Università Cattolica del Sacro Cuore},
            addressline={Largo Francesco Vito, 1}, 
            city={Roma},
            postcode={00168}, 
            country={Italy}}

\affiliation[12]{organization={Italian Society for Artificial Intelligence in Medicine (SIIAM)},
            city={Roma},
            country={Italy}}

\affiliation[13]{organization={Mayo Clinic},
            addressline={200 First Street SW}, 
            city={Rochester},
            postcode={55905}, 
            state={Minnesota},
            country={USA}}

\affiliation[14]{organization={Athinoula A. Martinos Center for Biomedical Imaging},
            addressline={149 13th Street}, 
            city={Charlestown},
            postcode={02129}, 
            state={Massachusetts},
            country={USA}}

\affiliation[15]{organization={Centro interdipartimentale di scienze mediche (CISMed),  Università degli studi di Trento},
            addressline={Via S. Maddalena, 1}, 
            city={Trento},
            postcode={38123}, 
            country={Italy}}

\affiliation[16]{organization={Sapienza Università di Roma},
            addressline={Piazzale Aldo Moro, 5}, 
            city={Roma},
            postcode={00185}, 
            country={Italy}}

\affiliation[17]{organization={Artificial Intelligence and Translational Imaging (ATI) Lab, Department of Radiology, School of Medicine, University of Crete},
            postcode={71003}, 
            country={Greece}}

\affiliation[18]{organization={Division of Radiology, Department of Clinical Science, Intervention and Technology (CLINTEC), Karolinska Institute},
            city={Huddinge},
            country={Sweden}}

\affiliation[19]{organization={Columbia University Medical Center},
            addressline={530 West 166th St}, 
            city={New York},
            postcode={10032}, 
            state={New York},
            country={USA}}

\affiliation[20]{organization={Department of Diagnostics and intervention, Diagnostic radiology, Umeå University},
            addressline={Universitetstorget, 4}, 
            city={Umeå},
            postcode={90187}, 
            country={Sweden}}

\cortext[1]{Corresponding author}

\begin{abstract}
Magnetic Resonance Imaging (MRI), Computed Tomography (CT), and Positron Emission Tomography (PET) provide complementary information about tissues. Medical image-to-image (I2I) translation enables virtual scanning by synthesizing a target modality from a source one without requiring an additional acquisition. Despite growing interest, many methods operate on 2D slices, are evaluated on isolated tasks under different experimental settings, and lack clinically oriented assessment. This work presents a reproducible benchmark for 3D I2I translation in oncological imaging that compares seven generative models: three Generative Adversarial Networks (Pix2Pix, CycleGAN, and SRGAN) and four latent models (Latent Diffusion Model, Latent Diffusion Model+ControlNet, Brownian Bridge, and Flow Matching). The benchmark comprises 77 experiments across eleven configurations drawn from five datasets, covering three anatomical regions (head/neck, lung, and pelvis) and four translation directions (cone-beam CT to CT, MRI to CT, CT to PET, and T2-weighted MRI to T2-FLAIR). Under the evaluated configurations, SRGAN achieves the highest quantitative image fidelity across all tasks, while latent models perform less well, due to information loss introduced by the variational autoencoder. A tumor-level analysis reveals that all models struggle with small lesions and that, in CT to PET synthesis, models reproduce tumor shape more reliably than tracer uptake values. A Visual Turing test involving 17 physicians, including 15 radiologists, shows near-chance classification accuracy (56.7\%), suggesting that experts struggle to distinguish real from synthetic volumes under the viewing conditions of the study. Expert preferences do not follow quantitative rankings, exposing a dissociation between quantitative metrics and clinical preference.
\end{abstract}



\maketitle

\section{Introduction}\label{sec:introduction}
Magnetic Resonance Imaging (MRI), Computed Tomography (CT), and Positron Emission Tomography (PET) are widely used medical imaging techniques that provide complementary information on tissues morphology and metabolic activity.
This complementarity is particularly useful in oncology, where images from different modalities help clinicians with cancer diagnosis, treatment planning, and response assessment \cite{dayarathna2024, doan2026}. 
As no single modality captures the full picture, oncological workflows routinely combine two or more modalities, but this places a considerable burden on both patients and healthcare systems. 
From the patient's perspective, undergoing multiple examinations increases risk due to cumulative exposure to ionizing radiation, contrast agents, or radio-tracers. 
For healthcare systems, the burden varies with the economic resources and infrastructure of the country involved.
While high-income countries often face the overuse of imaging, resulting in increased costs and pressure on radiology services, many low- and middle-income countries still lack adequate access to scanners and trained personnel \cite{kjelle2024}. 
This dual imbalance results either in unnecessary examinations and longer waiting lists, or in delayed diagnosis and treatment, ultimately leading to substantial disparities in cancer care \cite{who2024}.

Against this background, recent advances in generative Artificial Intelligence (AI) have facilitated progress in Image-to-Image (I2I) translation, a paradigm in which a model learns to transform an image from a source domain into a corresponding image in a target domain. 
In medical imaging, I2I translation provides the methodological foundation for {\em virtual scanning}, namely the synthesis of a target imaging modality from an already available source modality without requiring an additional acquisition. 
Although I2I approaches are advancing rapidly, they still face three major limitations that hinder their clinical applicability.
The first concerns data dimensionality: although MRI, CT, and PET are three-dimensional scans, most existing I2I methods are two-dimensional and operate on individual slices rather than full acquisition volumes \cite{sherwani2024}. 
While computationally efficient, two-dimensional processing neglects inter-slice context, with discontinuities that compromise anatomical coherence, essential for clinically reliable synthesis. 
Three-dimensional approaches are more effective at preserving spatial consistency and capturing long-range anatomical dependencies. However, their adoption is limited by the high computational and memory costs of processing high-resolution medical volumes.
The second limitation lies in the scarcity of studies addressing inter-modality translation. 
Unlike intra-modality synthesis, where the source and target images share similar acquisition, inter-modality tasks require the generative models to overcome significant differences in imaging physics, contrast mechanisms and semantic content \cite{dayarathna2024}. 
This makes cross-modality synthesis more challenging, but also more clinically valuable, since it can replace an acquisition rather than refine one.
The third limitation relates to evaluation practices \cite{fu2026}: current methods are typically evaluated using a narrow range of datasets, which limits clarity on how well they generalize when applied to different patient populations, anatomical regions, and translation tasks. 
Moreover, experimental protocols are often not sufficiently specified, thereby limiting reproducibility and rigorous cross-study comparisons. 
These shortcomings are particularly significant in oncology, where I2I approaches must do more than generating visually plausible images. 
They must also preserve clinically meaningful information, such as tumor conspicuity, in the regions that drive clinical decisions. 
Addressing these limitations is essential to move virtual scanning beyond proof-of-concept demonstrations and towards robust tools for clinical practice. 
Such tools align with the broader vision of {\em augmented intelligence}, in which AI is designed to support, rather than replace, clinical expertise \cite{rofena2025}.

Motivated by this need, this work explores I2I translation methods in oncological imaging to determine which generative AI architectures best preserve tumor fidelity and to what extent quantitative metrics commonly used in computer vision to assess image synthesis align with clinical perception.
Building on this rationale, we systematically searched medical imaging repositories for datasets that were both freely accessible and composed of oncological cases, in line with the clinical focus of this study.
The five selected datasets cover three anatomical regions (head and neck, lung, and pelvis) and multiple imaging modality pairs, providing a basis for benchmarking state-of-the-art generative AI models on I2I translation tasks.
Against these datasets, we propose a reproducible and standardized comparative evaluation of 3D I2I translation methods spanning two architectural families: Generative Adversarial Networks (GANs) and diffusion-based generative models. 
The models were selected to cover the methods most widely adopted for medical I2I translation, from the adversarial architectures that serve as baselines in the majority of medical I2I studies~\cite{dayarathna2024} to the diffusion-based approaches that now define the state of the art~\cite{kazerouni2023}, together with more recent solutions that have gained traction in medical benchmarks~\cite{valls2026,yazdani2026}.
The main contributions of the work can be summarized as following:

\begin{itemize}
    \item[C1)] We conduct a comparative analysis of seven state-of-the-art AI generative models for medical 3D I2I translation across five datasets, focusing on three anatomical regions: head/neck, lung, and pelvis; and four translation directions (cone-beam CT to CT, MRI to CT, CT to PET, MRI T2-weighted to T2-FLAIR), for a total of 77 experiments under shared training, inference, and evaluation.
    \item[C2)] We evaluate the generated images using metrics that assess image quality and structural fidelity.
    We focus on tumor regions to assess how pathological features are preserved in the generated images, categorizing the results according to lesion size and translation task, to identify failure modes that image-level metrics might miss.
    \item[C3)] We develop a web-based Visual Turing test platform and conduct a reader study with 17 physicians, including 15 radiologists, to assess whether generated volumes are perceptually indistinguishable from real scans and quantify the extent to which standard image quality metrics diverge from clinical preferences.
\end{itemize}

\section{Related works}\label{sec:related_works}

Over the past decade, GANs have played a central role in medical I2I translation, thanks to their ability to generate high-fidelity images while preserving structural information \cite{kazeminia2020}.
More recently, diffusion-based models have emerged as a promising alternative, offering greater sample diversity and more stable training, despite requiring higher computational costs \cite{kazerouni2023}.
The remainder of this section explores their application in medical I2I translation.

\textbf{GANs for Medical Image Translation.} 
2D slice-wise approaches dominate GAN-based medical I2I translation. 
The two prevailing paradigms are
Pix2Pix \cite{isola2017} and CycleGAN \cite{zhu2017}, which address paired and unpaired settings, respectively.
Numerous studies apply them to cross-modality synthesis such as MRI-to-CT \cite{nie2018, wolterink2017}, CT-to-MRI \cite{liu2020}, and multi-contrast MRI generation \cite{chartsias2018,dar2019}, processing each axial cross-section independently. 
Despite their effectiveness, 2D approaches ignore inter-slice spatial dependencies and produce spatial discontinuities. 
Since most clinical imaging modalities produce inherently 3D data, several studies have extended GAN architectures to process entire volumes directly, aiming to better preserve anatomical coherence.
EA-GAN \cite{yu2019} combines 3D convolutions with edge-aware learning to better preserve fine structural boundaries, while 3D-MedTranCSGAN \cite{poonkodi2023} uses a cascade of 3D-U-Blocks to iteratively refine the full volume through successive encoding–decoding pairs.
These approaches process entire volumes in a single forward pass, thereby capturing global context.
However, extending GAN architectures to 3D introduces computational costs due to the cubic growth in memory and operations, as well as challenges related to information compression, as encoders must distill volumes into compact latent representations without losing clinically relevant detail.
A solution is dividing the volume into patches that are processed independently and reassembled at inference time.
Guarrasi et al. \cite{guarrasi2025} train 3D variants of Pix2Pix and CycleGAN on volumetric patches processed independently and reconstruct the full volume at inference using a sliding window with overlap averaging.
Beyond memory considerations, most existing methods operate at a single spatial resolution and thus fail to leverage the hierarchical, multi-scale structure of medical volumetric data.
Ha et al. \cite{ha2025} propose SRGAN, a multi-resolution 3D GAN framework in which both generator and discriminator operate across multiple spatial scales.
The generator adopts a Dense-Attention UNet architecture, combining residual dense blocks for hierarchical feature extraction with attention mechanisms to fuse information across resolutions. 
The discriminator, also UNet-based, performs voxel-wise classification at each scale.
Experiments across diverse tasks (e.g., MRI-to-CT) and body regions (brain and pelvis) confirm the effectiveness of this framework, which outperforms EA-GAN in translation quality.

\textbf{From denoising diffusion to latent image translation.}
While being an effective paradigm for I2I translation in medical imaging, GAN-based methods still face well-documented limitations, like training instability, mode collapse, and sensitivity to architectural and hyperparameter choices that can degrade output diversity and reliability \cite{dhariwal2021}. 
This has motivated the exploration of alternative generative frameworks, among which Denoising Diffusion Probabilistic Models (DDPMs, \cite{ho2020}) have recently emerged as a promising direction.
DDPMs generate samples by progressively transforming noise into data through a neural network trained to reverse a Markovian noising process step by step.
Unconditional DDPMs generate samples from pure noise without conditioning on external information.
For I2I translation, however, the network must incorporate information from the source modality to produce a synthetic output that preserves anatomically coherence.
Conditional DDPMs achieve this by providing the source image as an additional network input, guiding the progressive generation of the image in the target modality.
Iterative denoising in DDPMs introduces a practical bottleneck: generating an image requires multiple network evaluations, making sampling more expensive than single-pass GAN inference, particularly when moving from 2D slices to full 3D volumes.
Whether this overhead translates into better synthesis quality remains task-dependent.
Moschetto et al. \cite{moschetto2025} investigate this through the first unified benchmark of GANs, conditional DDPMs, and flow matching models for 2D intra-MRI translation across three public brain imaging datasets.
Their results show that Pix2Pix outperforms all diffusion and flow variants in both structural fidelity and inference speed, a gap the authors attribute to the limited size of the 2D datasets used, insufficient for diffusion and flow models to learn the target distribution effectively \cite{bertrand2025, akbar2025}. 
In volumetric settings, where inter-slice consistency and long-range spatial context become critical, diffusion models narrow the gap with GANs and, in several tasks, surpass them.
Pan et al. \cite{pan2024} propose MC-IDDPM, the first conditional improved DDPM for 3D MRI-to-CT synthesis. 
On private brain and prostate datasets, MC-IDDPM outperforms GAN-based and standard DDPM baselines across all image-quality metrics, though inference remains orders of magnitude slower than single-pass alternatives.
The cubic memory cost associated with voxel-space diffusion has driven two alternatives: 2D slice-based architectures augmented with inter-slice consistency mechanisms \cite{choo2024, zhu2025}; and 3D Latent Diffusion Models (LDMs), which mitigate memory demands by operating in a compressed latent space instead of the original voxel domain.
A notable example of the latter is ALDM \cite{kim2024}, a 3D LDM which performs denoising in a VQGAN-compressed latent space \cite{esser2021}, enabling full-volume MRI translation without patch cropping.
On BraTS2021, ALDM surpasses all 2D and 3D GAN baselines and produces artifact-free sagittal views that slice-wise methods fail to preserve.
Sargood et al. \cite{sargood2026} extend latent diffusion to cross-modality tasks with CoCoLIT, a ControlNet-conditioned \cite{zhang2023} framework for 3D MRI-to-amyloid-PET synthesis.
On access-restricted brain datasets, CoCoLIT outperforms all baselines in both image fidelity and amyloid-positivity classification, indicating that latent diffusion can preserve clinically meaningful pathological information in the synthesized images.

\textbf{Clinical and oncological relevance.}
These studies reflect a progression from 2D slice-wise approaches to 3D solutions, which eliminate inter-slice artifacts but at a higher computational cost. 
Although a 3D generative model can reproduce the statistical distribution of a target modality with high fidelity, the synthesized image must preserve the underlying anatomical and functional information, a requirement that becomes critical in inter-modality I2I translation.
Pan et al. \cite{pan2024} validate MRI-to-CT synthesis using MC-IDDPM for brain and prostate tumors, showing that dose distributions computed on generated CT scans match those from planning CTs within clinically acceptable margins.
Graf et al. \cite{graf2023} demonstrate that 3D Pix2Pix and Denoising Diffusion Implicit Models (DDIM) for MRI-to-CT translation of the spine produce segmentation masks accurate enough to support biomechanical modeling, a prerequisite for surgical planning and vertebral fracture assessment in metastatic disease.
The translation problem changes when the target modality is functional rather than anatomical, like in MRI-to-PET and CT-to-PET synthesis.
Rajagopal et al. \cite{rajagopal2023} show that synthetic FDG-PET generated from whole-body MRI can replace real PET sinograms in SUV quantification experiments with synthetic lesion insertion, supporting the development and qualification of PET/MRI reconstruction algorithms without requiring new patient acquisitions.
Sargood et al. \cite{sargood2026} demonstrate that latent diffusion can synthesize amyloid-PET scans from structural MRI with sufficient accuracy to classify amyloid positivity in Alzheimer's disease screening, a task with direct implications for early oncological differential diagnosis of brain cancer.
Guarrasi et al. \cite{guarrasi2025} address CT-to-PET translation by segmenting whole-body CT volumes into anatomical regions and training region-specific Pix2Pix, an approach that acknowledges the heterogeneity of tracer bio-distribution across body regions.

\textbf{Open Challenges in Medical I2I Translation.}
Despite these advances, there are still four issues with medical imaging translation.
First, not all I2I tasks carry equal oncological relevance. MRI-to-CT synthesis for radiotherapy planning and MRI/CT-to-PET translation for staging and response monitoring represent the most clinically motivated directions, while intra-modal MRI and CT synthesis serve complementary but narrower roles \cite{bahloul2024}.
Second, a pronounced anatomical bias persists, with most studies focusing on the brain, followed by the spine and lung. 
Whole-body approaches either divide the anatomy into separate regions or highlight the difficulty of capturing anatomical heterogeneity within a single model \cite{guarrasi2025}.
Third, dataset availability further constrains progress, with reproducibility limited by the predominance of private institutional data~\cite{friedrich2024}. 
Only a few studies rely on multiple public benchmarks, and initiatives such as the SynthRAD challenge \cite{synthrad2025} represent a step toward standardized evaluation, but broader multi-site, multi-task datasets with clinical endpoint validation remain a critical gap for translating these methods into oncological practice.
Finally, generalization remains limited, as few studies attempt multiple translation tasks using a single model, and evidence of consistent accuracy across inter-modality pairs remains scarce.
The novelty of this work lies in addressing this gap through a comparative evaluation of 3D I2I translation methods across different anatomical regions and translation directions to assess their generalizability across anatomical and imaging contexts.

\section{Materials and Methods}\label{sec:materials_methods}

\subsection{Experimental Setup}\label{sec:datasets}
This work establishes a standardized benchmark for 3D I2I translation in oncological imaging. 
To construct this benchmark, we conducted a systematic search of publicly available medical imaging repositories, selecting those that were freely accessible and composed of oncological cases.
Five datasets satisfied both criteria: SynthRAD23~\cite{synthrad2023}, SynthRAD25~\cite{synthrad2025}, BraTS23~\cite{brats2023}, autoPET~\cite{gatidis2022}, and EnhancePET~\cite{ferrara2026}.
These collections span two inter-modality and two intra-modality translation directions, providing a clinically grounded basis for benchmarking generative models, and three anatomical regions (lung, head/neck, and pelvis), that account for a significant proportion of cancer incidence and mortality worldwide \cite{bray2024, siegel2026}.
Tumors primarily diagnosed through 2D imaging, such as breast cancer (mammography) and skin cancer (dermoscopy), fall outside the scope of this work, although emerging 3D modalities such as breast MRI represent a natural future extension.

\begin{figure}
  \centering
    \includegraphics[width=\textwidth]{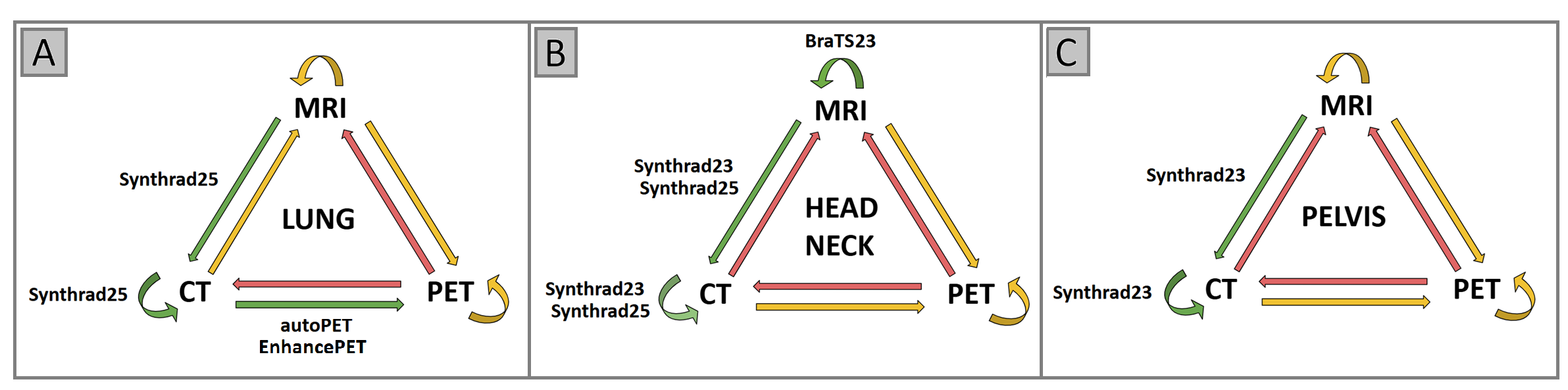}
    \caption{Overview of paired I2I translation tasks selected for this study, grouped by anatomical region (lung, \textbf{A}; head/neck, \textbf{B}; and pelvis, \textbf{C}). Triangle vertices represent the three imaging modalities (CT, MRI, and PET). Inter-modality translations are represented by arrows between vertices, while intra-modality tasks are indicated by self-loops. Arrow colors are assigned based on clinical relevance and public dataset availability: green arrows denote tasks included in this benchmark, with the corresponding dataset name reported alongside; yellow arrows indicate clinically relevant directions for which no publicly accessible paired dataset was identified; red arrows denote synthesis directions considered clinically not relevant in the corresponding anatomical context.}\label{fig1}
\end{figure}

Figure~\ref{fig1} illustrates the imaging modalities and I2I translation tasks for the selected body regions. 
The triangle’s vertices represent CT, MRI, and PET, while its edges connect modality pairs, with arrows indicating the source-to-target direction.
Not all directions are equally relevant: we consider an I2I task clinically relevant when acquiring the target modality entails additional cost, radiation exposure, or logistical burden for the patient.
Conversely, we consider a direction not clinically meaningful when the target is the principal imaging modality for the selected anatomical region and therefore routinely acquired (like MRI for pelvis and head/neck), or when the source lacks the structural information to reliably infer the target (e.g., PET-to-CT or PET-to-MRI). 
Accordingly, we classify each direction along an edge as a clinically relevant task with publicly available paired data, suitable for inclusion in this benchmark (green arrows), a clinically relevant task currently lacking such data (yellow arrows),  or a clinically not relevant direction (red arrows). 
We restrict our evaluation to the first group, for which publicly available paired data enable reproducible comparisons across methods.
The dataset used for each task is indicated alongside the corresponding arrow.

The SynthRAD23 dataset, provided by the SynthRAD Grand Challenge organizers~\cite{synthrad2023}, covers two tasks on the head/neck and pelvic regions, yielding four experimental configurations in total. 
The first task consists of MR/CT image pairs and addresses T1-weighted (T1w) MRI-to-CT synthesis.
T1w MRI provides high soft-tissue contrast, facilitating the delineation of tumor boundaries and surrounding anatomy. CT, although required for radiotherapy dose planning, provides less soft-tissue contrast. 
Synthesizing CT from MRI therefore has the potential to remove the need for a separate CT acquisition in the radiotherapy planning workflow, reducing radiation exposure while preserving the anatomical detail needed for accurate dose calculation.
The second task consists of Cone Beam CT (CBCT)/CT image pairs and focuses on CBCT-to-CT translation.
CBCT is acquired at treatment time and exposes patients to a reduced dose compared to CT, but this comes at the cost of increased noise, artifacts, and reduced contrast, which limits its reliability for precise dose recalculation. 
Translating CBCT into CT-like quality images could enable accurate imaging for adaptive radiotherapy, without requiring the acquisition of an additional CT scan.

The SynthRAD25 dataset is the second edition of the SynthRAD Grand Challenge, extending its predecessor to three anatomical regions ~\cite{synthrad2025}: head/neck, thorax, and abdomen. 
As in SynthRAD23, two synthesis tasks are defined for each region: T1w MRI-to-CT and CBCT-to-CT. 
For this benchmark, only two of the three anatomical regions were retained, to maintain consistency with the other datasets, which cover the brain, the pelvis, and the lungs. 
For both SynthRAD datasets, we excluded the official challenge validation and test sets because their ground-truth CT scans were unavailable at the time of this study. 
We used only the challenge training cases, for which ground-truth scans were available, and divided them into training (75\%) and test (25\%) sets independently for each task–anatomy combination. 
Table~\ref{tbl1} reports the resulting number of cases in each subset.

\begin{table}[t!]
\caption{Overview of the datasets used in this study. Each row specifies the I2I translation task, the representative downstream application, the selected dataset, the anatomical region, and the total number of patients available to train the I2I model, with the test set size in parentheses. }\label{tbl1}
\begin{tabular*}{\tblwidth}{cm{5.5cm}ccc}
\toprule
 \makecell[c]{\textbf{Task} \\ \textbf{(Source-to-Target)}} & \textbf{Applications} & \textbf{Dataset} & \textbf{District} & \textbf{Patients (Test)}\\
\midrule
\multirow{4}{*}{MRI T1w-to-CT}  & \multirow{4}{=}{MRI-only radiotherapy planning: eliminate CT acquisition for dose calculation in brain, head/neck, and pelvic cancers}  & \multirow{2}{*}{Synthrad23 \cite{synthrad2023}} & Pelvis      & 180 (45)   \\
&                    &                               & Head/Neck     & 180 (45)   \\
&                    & \multirow{2}{*}{Synthrad25 \cite{synthrad2025}} & Lung       & 182 (46)   \\
&                    &                               & Head/Neck & 156 (39)   \\
\midrule
\multirow{4}{*}{CBCT-to-CT} & \multirow{4}{=}{Online adaptive radiotherapy: enable dose recalculation from daily CBCT without additional full-dose CT}  & \multirow{2}{*}{Synthrad23 \cite{synthrad2023}} & Pelvis      & 180 (45)   \\
&                    &                               & Head/Neck     & 180 (45)   \\
&                    & \multirow{2}{*}{Synthrad25 \cite{synthrad2025}} & Lung       & 258 (65)   \\
&                    &                               & Head/Neck & 260 (65)   \\
\midrule
MRI T2w-to-T2f                &     Completion of the MRI protocol: recover missing FLAIR sequences due to acquisition failures or time constraints in brain tumor imaging          & BraTS23 \cite{brats2023}                    & Brain      & 1251 (313) \\
\midrule
\multirow{2}{*}{CT-to-PET}  & \multirow{2}{=}{Reduce radiotracer exposure in lung cancer follow-up}  & autoPET \cite{gatidis2022}                       & \multirow{2}{*}{Lung} & 150 (38)   \\
&                    & EnhancePET \cite{ferrara2026}              &                       & 583 (146)  \\
                                        
\bottomrule
\end{tabular*}
\end{table}

The BraTS23 dataset originates from the ASNR-MICCAI Brain Tumour Segmentation 2023 Glioma Challenge~\cite{brats2023} and is used in this benchmark for intra-MRI translation between T2-weighted (T2w) and T2-FLAIR (T2f).
T2f suppresses the signal from cerebrospinal fluid, thereby improving the detection of tumors that may otherwise be hard to identify, and is the reference sequence for determining tumor margins and peri-ventricular lesions.
Synthesizing T2f from T2w images can help recover contrast relevant to lesion assessment when the T2f sequence is missing, corrupted, or excluded from the protocol to reduce acquisition time.
For this study, we used the training set of the challenge, comprising 1,251 cases with publicly available ground-truth segmentation labels, divided into 938 training and 313 test cases (Table \ref{tbl1}), using the same 75\%–25\% train–test split applied to the SynthRAD datasets.

The autoPET dataset is derived from a publicly available whole-body FDG-PET/CT collection~\cite{gatidis2022} that includes patients diagnosed with malignant lymphoma, melanoma, and lung cancer, as well as negative-control studies of subjects without confirmed malignancy. 
To maintain consistency with the other datasets, only lung cancer cases were retained; subjects with lymphoma, melanoma, or no confirmed malignancy were excluded, yielding a final subset of 150 paired CT/FDG-PET cases. 
The most widely used PET tracer in oncology is \textsuperscript{18}F-fluorodeoxyglucose (FDG), a glucose analogue that accumulates more in metabolically active tissues. 
As malignant tumors typically exhibit increased glucose uptake, 
FDG-PET provides functional information that complements the anatomical detail of CT or MRI. However, PET acquisition requires radio-tracer administration, additional scan time, and specialized infrastructure. CT-to-PET translation aims to approximate this functional information from a lower-cost CT scan, with the potential to improve access to PET-like imaging in resource-limited settings.

Finally, the EnhancePET dataset is a large multi-center whole-body FDG-PET/CT collection~\cite{ferrara2026} hosted on the AWS Open Data Registry, comprising approximately 1,600 subjects. For this study, the 1,014 subjects already covered in the autoPET cohort were excluded to avoid overlap between the datasets. 
The remaining subjects, all diagnosed with lung cancer, yielded a final cohort of 583 CT/FDG-PET pairs.

For each dataset, pre-processing builds on the steps provided with the dataset release, followed by a pipeline comprising background masking, resampling, modality-specific intensity clipping, normalization, spatial padding, and computation of a shared source–target foreground mask (see Figure~\ref{fig:benchmark}, pre-processing module). 
Paired 3D patches of size $96 \times 96 \times 96$ voxels are then extracted from the processed volumes for model training.
Details of the dataset-specific procedures and the pre-processing pipeline are provided in the Supplementary Material, Section~\ref{app:datasets}.

\subsection{Models Architectures and Training}\label{sec:models}

\begin{figure}
  \centering
    \includegraphics[width=\textwidth]{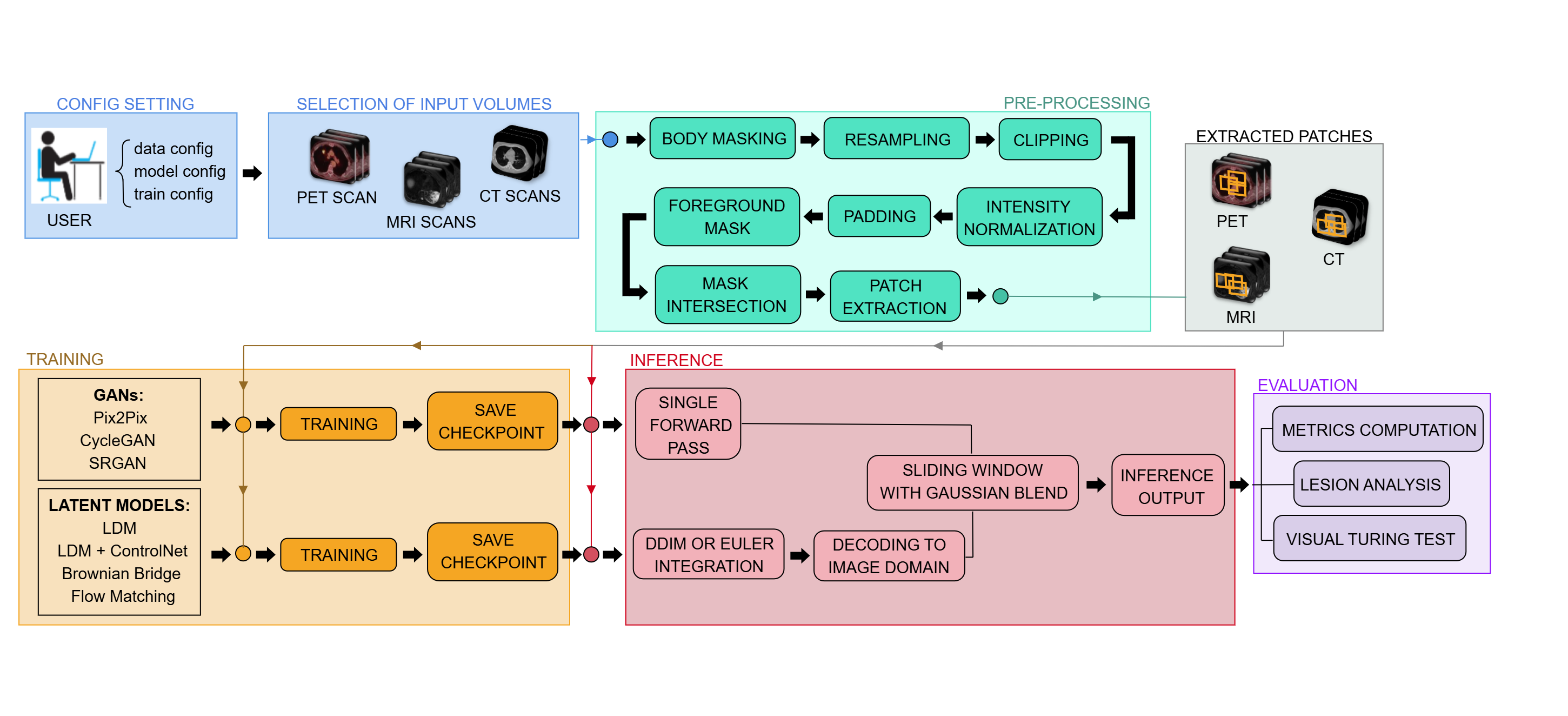}
    \caption{The pipeline presented in this work consists of five stages: (1) a configuration module, where the user specifies data, model, and processing parameters, according to which the training and test sets are defined; (2) a pre-processing module, which applies a sequence of pre-processing steps to produce paired source–target volume patches; (3) a training stage: GAN-based models are trained in the image domain and latent generative models on VAE-encoded representations; (4) an inference stage, where trained models generate full-resolution patches via model-specific inference strategies (single forward pass for GANs, DDIM sampling for diffusion models, Euler integration for Flow Matching), followed by stiching of the generated patches; (5) an evaluation module, in which generated volumes are used to compute whole-volume metrics, perform lesion-stratified analysis and conduct the clinician Visual Turing test.
} \label{fig:benchmark}
\end{figure}

This work benchmarks seven state-of-the-art AI generative models for paired medical image synthesis, spanning two architectural families: GANs and diffusion-based models. 
Pix2Pix and CycleGAN serve as GAN baselines in the majority of medical I2I studies~\cite{dayarathna2024}; SRGAN has recently been proposed as a multi-resolution 3D GAN framework in which both generator and discriminator operate across multiple spatial scales. 
Among latent generative models, LDM-based approaches, with and without ControlNet, represent the current standard for diffusion-based medical synthesis~\cite{kazerouni2023}.
The Brownian Bridge model has drawn growing interest for MRI translation tasks~\cite{valls2026}, and flow matching has recently emerged as a competitive alternative in medical imaging benchmarks~\cite{yazdani2026}.

The three GAN models operate directly on the $96\times96\times96$ voxel patches and follow the standard adversarial training paradigm, in
which a generator and a discriminator are optimized simultaneously.
Within this family, Pix2Pix learns a source-to-target mapping through adversarial and paired reconstruction losses; CycleGAN introduces bidirectional mappings constrained by cycle consistency; and SRGAN uses generator and discriminator architectures that operate across multiple spatial scales.
Conversely, the four latent generative models operate in the compressed latent space of a shared Variational AutoEncoder (VAE), pre-trained independently for each dataset and kept frozen during downstream training.
The VAE encodes input patches into $24\times24\times24$ latent volumes, reducing each spatial dimension by a factor of four.
Within the shared latent representation, the models differ in how they formulate the source-to-target translation.
LDM-Palette (with and without ControlNet conditioning) learns to iteratively denoise a target latent conditioned on the source, using a DDPM noise schedule during training and DDIM sampling at inference.
BBridge instead replaces the Gaussian noise prior with a Brownian Bridge process that interpolates directly between source and target latent vectors, removing the need to start from pure noise. 
Flow Matching learns a deterministic velocity field that transports the source latent to the target latent along straight-line trajectories via numerical ODE integration.

As reported in Table~\ref{tbl1}, patients from each dataset have been divided into training and test sets using a 75\%–25\% split. 
Five patients from each training partition were then held out for validation, and the same validation subset has been used across all tasks.
All models were trained for 1000 epochs on a single NVIDIA A40 GPU (48GB), using a batch size of three randomly sampled $96\times96\times96$ voxel patches per iteration, following~\cite{ha2025}. No model underwent hyperparameter tuning, as dataset-specific optimization was beyond the scope of this benchmark. Instead, we used fixed hyperparameter configurations to ensure a consistent experimental protocol across datasets. Full architectural details are provided in the Supplementary Material, Section~\ref{app:implementation}, and training hyperparameters are summarized in Table~\ref{tab2}.

\begin{table}[t]
\centering
\caption{Training hyperparameters for all benchmarked models.}
\label{tab2}
\begin{tabular}{lccccc}
\toprule
Model & Optimiser & LR & Schedule & $\beta_1,\beta_2$ & Epochs \\
\midrule
Pix2Pix        & Adam  & $10^{-4}$          & step $\times0.5$ @ 200/400/600/800 & 0.5, 0.999 & 1000 \\
CycleGAN       & Adam  & $10^{-4}$          & step $\times0.5$ @ 200/400/600/800 & 0.5, 0.999 & 1000 \\
SRGAN          & Adam  & $10^{-4}$          & MultiStepLR $\times0.5$            & 0.9, 0.999 & 1000 \\
VAE            & Adam  & $10^{-4}$          & warm-up (10 ep, $\times0.1$)        & 0.9, 0.999 & 1000 \\
LDM-Palette    & AdamW & $1.5\times10^{-5}$ & ReduceLROnPlateau ($\times0.5$, 30 ep) & default & 1000 \\
Brownian Bridge& AdamW & $1.5\times10^{-5}$ & ReduceLROnPlateau ($\times0.5$, 3k it) & default & 1000 \\
ControlNet     & AdamW & $1.5\times10^{-5}$ & Cosine annealing ($T_\mathrm{max}=1000$) & default & 1000 \\
Flow Matching  & AdamW & $1.5\times10^{-5}$ & ReduceLROnPlateau ($\times0.5$, 30 ep) & default & 1000 \\
\bottomrule
\end{tabular}
\end{table}

Beyond the individual models described above, a central contribution of this work is the benchmarking framework, which we release as an open-source repository to serve as a reusable and extensible research instrument for the medical image synthesis community. 
To the best of our knowledge, this is the first reproducible 3D benchmark to unify pre-processing, training, inference, and evaluation across heterogeneous oncological translation tasks, spanning 4 translation directions, 3 body regions, and 11 dataset configurations within a single pipeline.
Specifically, this benchmark extends the line of investigation of \cite{moschetto2025} along four dimensions: (i) dimensionality (3D volumetric vs.\ 2D slice-wise processing); (ii) scope (4 translation directions across 3 body regions vs.\ a single intra-MRI task); (iii) scale (77 experiments across 11 dataset configurations vs.\ 3 datasets); and (iv) evaluation depth (tumor-level analysis and a Visual Turing test vs.\ standard image quality metrics only).
To support these comparisons, the framework uses a modular design that separates the generative model from the surrounding pipeline (Figure~\ref{fig:benchmark}), so that new architectures use the same pre-processing, patch sampling, training infrastructure, inference procedure, and evaluation tools as the existing baselines.
Designed in this way, the framework allows a more controlled comparison of the generative models.

\subsection{Evaluation Metrics}\label{sec:metrics}

Before evaluation, full volumes are reconstructed using a sliding-window inference shared by all models, with patches extracted at an overlap fraction of 0.625 (Figure~\ref{fig:benchmark}, inference stage).
GANs translate each patch in a single forward pass, whereas latent models use iterative sampling followed by VAE decoding.
The predicted patches are then combined using Gaussian blending with a sigma scale of 0.125, which assigns greater weight to voxels in the central region of each patch and progressively less weight towards its boundaries.
We evaluate the resulting volumes against the corresponding ground-truth images using Peak Signal-to-Noise Ratio (PSNR) and Structural Similarity Index Measure (SSIM)~\cite{difeola2023}.
For the autoPET and BraTS23 datasets, for which tumor masks are available, the same metrics are computed within lesion regions and stratified according to lesion size.
PSNR and SSIM do not fully capture perceptual image quality~\cite{roberts2023,tang2024}.
For this reason, we complement the quantitative evaluation with a Visual Turing test to assess image quality as perceived by medical experts, following previous studies~\cite{myong2023, jang2023, phelps2015, hoeijmakers2024,guarrasi2024}.
To the best of our knowledge, these studies have not examined the perceived quality of synthetic 3D medical images across both intra- and inter-modality translation tasks.

The Visual Turing test comprises 50 questions divided into three parts, with balanced coverage of all translation tasks and anatomical regions (Table~\ref{tbl3}). 
The test evaluates four generative models, selecting the two best-performing models from each architectural family based on quantitative metrics (see Section \ref{sec:perceptual}).
\textbf{Part~1} presents 15 single images (7 real and 8 synthetic) and asks participants to classify them as real or AI-generated, as a binary test following the paradigm used in prior studies~\cite{ myong2023,jang2023}. 
\textbf{Part~2} displays 18 side-by-side image pairs, and participants select the one they preferred for diagnostic purposes, or indicate that no meaningful difference could be identified.
We select this pairwise comparison design in place of a Likert-scale rating because it yields higher inter- and intra-observer agreement in subjective image quality assessment~\cite{tang2024,phelps2015, hoeijmakers2024}.
Both images in each pair are synthetic and always drawn from the same patient but generated by different models, so as to eliminate inter-patient variability from the comparison.
Three pairs of identical images are embedded as control questions to monitor response consistency.
\textbf{Part~3} presents 17 image triplets, and participants assign a rank from 1 (most realistic) to 3 (least realistic) to each one, providing an ordinal quality measure across models under matched conditions.
In each triplet, images are either real or generated by different models. As for Part 2, all images within a triplet belong to the same patient to eliminate inter-patient variability.
Two control questions are included in this part, each containing one real image and two outputs from the same model, to identify potential response bias when the identical outputs were ranked more than one position apart.

\begin{table}[t!]
\centering
\caption{Distribution of the 50 questions across the three parts of the visual Turing test, organized by translation task and anatomical region. Part~1 (real vs.\ synthetic identification, 15 questions), Part~2 (pairwise preference, 15+3 control questions), and Part~3 (quality rating, 15+2 control questions). Each part covers all four translation categories and three anatomical regions to ensure balanced task representation across evaluation modes. The pelvis region has fewer questions across all parts, as fewer translation tasks in the benchmark map to this anatomical region. Similarly, CBCT-to-CT is assigned fewer questions than the other translations, reflecting the narrower domain gap of this intra-modality task.}
\label{tbl3}
\resizebox{\textwidth}{!}{%
\begin{tabular}{lcccccccccccc}
\toprule
& \multicolumn{4}{c}{\textbf{Part 1}}
& \multicolumn{4}{c}{\textbf{Part 2}}
& \multicolumn{4}{c}{\textbf{Part 3}} \\
\cmidrule(lr){2-5} \cmidrule(lr){6-9} \cmidrule(lr){10-13}
\textbf{Translation}
& \textbf{Head/Neck} & \textbf{Pelvis} & \textbf{Lung} & \textbf{Total}
& \textbf{Head/Neck} & \textbf{Pelvis} & \textbf{Lung} & \textbf{Total}
& \textbf{Head/Neck} & \textbf{Pelvis} & \textbf{Lung} & \textbf{Total} \\
\midrule
T1w-to-CT
& 1 & 2 & 1 & \textbf{4}
& 1 & 2 & 1 & \textbf{4}
& 1 & 2 & 1 & \textbf{4} \\
CBCT-to-CT
& 1 & 1 & 1 & \textbf{3}
& 1 & 1 & 1 & \textbf{3}
& 1 & 1 & 1 & \textbf{3} \\
T2w-to-T2f
& 4 & -- & -- & \textbf{4}
& 4 & -- & -- & \textbf{4}
& 4 & -- & -- & \textbf{4} \\
CT-to-PET
& -- & -- & 4 & \textbf{4}
& -- & -- & 4 & \textbf{4}
& -- & -- & 4 & \textbf{4} \\
\midrule
Control
& -- & -- & -- & --
& 1 & 1 & 1 & \textbf{3}
& 1 & -- & 1 & \textbf{2} \\
\textbf{Total}
& \textbf{6} & \textbf{3} & \textbf{6} & \textbf{15}
& \textbf{7} & \textbf{4} & \textbf{7} & \textbf{18}
& \textbf{7} & \textbf{3} & \textbf{7} & \textbf{17} \\
\bottomrule
\end{tabular}%
}
\end{table}

We administered this test through a custom web-based platform with interactive volumetric visualization.
The platform supports NIfTI volumes, configurable question types, per-volume windowing controls, multi-planar navigation, and anonymous response collection.
For each question, the tool displays each volume using the axial, sagittal, and coronal views, alongside a 3D surface reconstruction rendered using NiiVue v0.50.0~\cite{niivue}. 
Adjustable brightness, contrast, and slice controls are provided for each volume (see Supplementary Material, Section~\ref{app:turing_platform} for more details). 
A total of 17 clinicians participated in the study: 15 radiologists and 2 physicians with a different clinical specialization, from 12 different institutions and 5 different countries. 
In terms of experience, 12 participants reported more than 10 years  of clinical practice, 3 between 5 and 10 years, and 2 less than 5 years. 
Regarding primary imaging modality experience, 8 reported working with CT and MRI,  4 with all three modalities, 2 with MRI only,  1 with CT only, 1 with PET only, and 1 with CT and PET. 
All responses were anonymous and no personal data was collected. 

To quantify uncertainty in the reader assessments, test whether responses differ from chance, and assess agreement among readers, we perform the following statistical analyses (results in Section~\ref{sec:perceptual}).
We report Part~1 classification accuracy overall and per translation task with 95\% confidence intervals obtained by bootstrap resampling of readers (5000 resamples), accounting for within-reader response dependence. 
Individual reader accuracies are compared with chance (50\%) using exact binomial tests with Holm--Bonferroni correction across readers, while differences between experience and subspecialty groups are assessed using Mann--Whitney $U$ tests. 
For Part~2, reader-level net preferences between model pairs are tested against zero using Wilcoxon signed-rank tests, with Holm--Bonferroni correction across comparisons and matched-pairs rank-biserial correlations ($r$) as effect sizes. 
The same approach is used in Part~3 to test reader-level pairwise ranking proportions against 0.5 and the proportion of real images ranked first against $1/3$. 
Inter-reader agreement is assessed using Fleiss' $\kappa$ for categorical responses in Parts~1 and~2 and Kendall's coefficient of concordance $W$, averaged across triplets, for rankings in Part~3. 
Control questions are excluded from all statistics.

\section{Results and Discussion}\label{sec:results}

We analyze the results at three complementary levels: quantitative performance, tumor-level fidelity, and perceptual realism.
Section~\ref{sec:quantitative} examines the performance of the selected models across modalities and anatomical regions, to determine whether a single generative framework performs consistently across tasks or whether model selection must be tailored to each translation.
Qualitative comparisons are provided in 
Supplementary Materials, Section~\ref{app:additional_results}.
Section~\ref{sec:tumor_level} extends the evaluation beyond whole-volume metrics to quantify fidelity within tumor regions, where synthesis errors may directly affect clinical interpretation.
Finally, Section~\ref{sec:perceptual} assesses perceptual realism by comparing quantitative rankings with human perception, evaluated through the Visual Turing test administered to the physicians.

\subsection{Quantitative Performance Across Modalities and Regions}\label{sec:quantitative}

\providecommand{\heatbar}{{\setlength{\fboxsep}{0pt}\setlength{\fboxrule}{0.3pt}%
  \raisebox{-0.1em}{\fbox{%
    \textcolor{hl!0}{\rule{0.9em}{0.8em}}\textcolor{hl!10}{\rule{0.9em}{0.8em}}%
    \textcolor{hl!20}{\rule{0.9em}{0.8em}}\textcolor{hl!30}{\rule{0.9em}{0.8em}}%
    \textcolor{hl!40}{\rule{0.9em}{0.8em}}\textcolor{hl!50}{\rule{0.9em}{0.8em}}%
    \textcolor{hl!60}{\rule{0.9em}{0.8em}}}}}}
\begin{table}[h]
\caption{PSNR (dB) and SSIM (mean with standard deviation as subscript) across anatomical districts, tasks, and
models. Bold and underlined values indicate best and second best result per row, respectively.
Within each row, cell shading encodes the gap to the best model on a logarithmic scale, from white (worst) to dark green (best).
The rightmost column reports the VAE reconstruction level, which represents a
reference for all latent-based architectures.  The VAE column is not shaded.}\label{tbl:generation}
\fontsize{8}{9.5}\selectfont
\setlength{\tabcolsep}{2pt}
\begin{tabular*}{\tblwidth}{@{\extracolsep{\fill}}lccccccccccc}
\toprule
 & \textbf{District} & \textbf{Task} & \textbf{Dataset}
  & \textbf{SRGAN} & \textbf{CycleGAN} & \textbf{Pix2Pix}
  & \textbf{BBridge} & \textbf{LDM} & \textbf{LDM+CN} & \textbf{FlowM}
  & \textbf{VAE} \\
\midrule

\multirow{11}{*}{\rotatebox[origin=c]{90}{PSNR $\uparrow$}}
& Brain & T2w-to-T2f & BraTS23
    & \cellcolor{hl!60}$\mathbf{25.28}_{2.90}$ & \cellcolor{hl!22}$23.52_{3.02}$ & \cellcolor{hl!27}$\underline{23.89}{}_{2.72}$
    & \cellcolor{hl!16}$23.00_{2.67}$ & \cellcolor{hl!0}$20.84_{1.97}$ & \cellcolor{hl!0}$20.83_{1.95}$ & \cellcolor{hl!22}$23.55_{2.56}$
    & $28.97_{1.55}$ \\
\cmidrule{2-12}
& \multirow{4}{*}{Head-Neck}
  & \multirow{2}{*}{T1w-to-CT} & Synthrad23
    & \cellcolor{hl!60}$\mathbf{27.48}_{1.47}$ & \cellcolor{hl!12}$24.77_{1.54}$ & \cellcolor{hl!0}$23.03_{0.95}$
    & \cellcolor{hl!14}$24.99_{0.85}$ & \cellcolor{hl!14}$24.96_{0.80}$ & \cellcolor{hl!17}$\underline{25.24}{}_{0.88}$ & \cellcolor{hl!13}$24.82_{0.83}$
    & $29.33_{0.88}$ \\
& & & Synthrad25
    & \cellcolor{hl!60}$\mathbf{26.32}_{3.01}$ & \cellcolor{hl!23}$\underline{24.45}{}_{2.39}$ & \cellcolor{hl!8}$22.77_{1.89}$
    & \cellcolor{hl!2}$21.78_{1.47}$ & \cellcolor{hl!4}$22.13_{1.50}$ & \cellcolor{hl!4}$22.18_{1.53}$ & \cellcolor{hl!0}$21.42_{1.45}$
    & $22.66_{1.77}$ \\
\cmidrule(lr){3-12}
& & \multirow{2}{*}{CBCT-to-CT} & Synthrad23
    & \cellcolor{hl!60}$\mathbf{28.88}_{2.05}$ & \cellcolor{hl!41}$\underline{28.21}{}_{2.60}$ & \cellcolor{hl!22}$26.98_{1.84}$
    & \cellcolor{hl!9}$25.43_{1.65}$ & \cellcolor{hl!0}$23.98_{2.20}$ & \cellcolor{hl!1}$24.25_{2.30}$ & \cellcolor{hl!11}$25.68_{1.84}$
    & $26.66_{1.35}$ \\
& & & Synthrad25
    & \cellcolor{hl!60}$\mathbf{29.50}_{1.70}$ & \cellcolor{hl!29}$\underline{28.65}{}_{1.66}$ & \cellcolor{hl!6}$27.10_{1.53}$
    & \cellcolor{hl!0}$26.45_{1.53}$ & \cellcolor{hl!1}$26.54_{1.77}$ & \cellcolor{hl!2}$26.66_{1.77}$ & \cellcolor{hl!1}$26.55_{1.92}$
    & $29.61_{2.19}$ \\
\cmidrule{2-12}
& \multirow{4}{*}{Lung}
  & \multirow{2}{*}{CT-to-PET} & autoPET
    & \cellcolor{hl!60}$\mathbf{34.21}_{3.19}$ & \cellcolor{hl!42}$\underline{33.41}{}_{2.84}$ & \cellcolor{hl!22}$31.68_{2.68}$
    & \cellcolor{hl!0}$27.78_{1.16}$ & \cellcolor{hl!2}$28.33_{1.26}$ & \cellcolor{hl!3}$28.50_{1.50}$ & \cellcolor{hl!0}$27.89_{1.11}$
    & $35.10_{1.78}$ \\
& & & EnhancePET
    & \cellcolor{hl!60}$\mathbf{42.77}_{7.53}$ & \cellcolor{hl!37}$\underline{41.21}{}_{7.41}$ & \cellcolor{hl!37}$41.16_{6.84}$
    & \cellcolor{hl!0}$33.71_{3.09}$ & \cellcolor{hl!0}$33.79_{3.25}$ & \cellcolor{hl!6}$35.77_{2.63}$ & \cellcolor{hl!6}$35.63_{2.38}$
    & $39.46_{4.78}$ \\
\cmidrule(lr){3-12}
& & T1w-to-CT & Synthrad25
    & \cellcolor{hl!60}$\mathbf{25.74}_{1.51}$ & \cellcolor{hl!29}$\underline{24.39}{}_{1.67}$ & \cellcolor{hl!16}$23.17_{1.84}$
    & \cellcolor{hl!11}$22.68_{1.46}$ & \cellcolor{hl!0}$20.88_{1.64}$ & \cellcolor{hl!0}$20.87_{1.76}$ & \cellcolor{hl!6}$21.83_{1.82}$
    & $26.84_{3.01}$ \\
& & CBCT-to-CT & Synthrad25
    & \cellcolor{hl!60}$\mathbf{28.07}_{1.98}$ & \cellcolor{hl!37}$\underline{27.49}{}_{1.95}$ & \cellcolor{hl!18}$26.54_{1.76}$
    & \cellcolor{hl!0}$24.96_{1.97}$ & \cellcolor{hl!1}$24.97_{2.01}$ & \cellcolor{hl!1}$25.07_{2.03}$ & \cellcolor{hl!0}$24.90_{2.12}$
    & $26.35_{2.56}$ \\
\cmidrule{2-12}
& \multirow{2}{*}{Pelvis}
  & T1w-to-CT & Synthrad23
    & \cellcolor{hl!60}$\mathbf{29.15}_{1.64}$ & \cellcolor{hl!22}$27.04_{1.77}$ & \cellcolor{hl!27}$\underline{27.48}{}_{1.95}$
    & \cellcolor{hl!13}$26.08_{1.44}$ & \cellcolor{hl!0}$23.88_{2.11}$ & \cellcolor{hl!2}$24.25_{2.08}$ & \cellcolor{hl!12}$25.94_{1.37}$
    & $31.07_{1.57}$ \\
& & CBCT-to-CT & Synthrad23
    & \cellcolor{hl!60}$\mathbf{29.83}_{2.29}$ & \cellcolor{hl!35}$\underline{28.61}{}_{2.72}$ & \cellcolor{hl!32}$28.40_{1.60}$
    & \cellcolor{hl!5}$24.84_{1.16}$ & \cellcolor{hl!0}$23.76_{1.30}$ & \cellcolor{hl!0}$23.73_{1.47}$ & \cellcolor{hl!5}$24.78_{1.21}$
    & $24.22_{3.58}$ \\

\midrule
\midrule
\multirow{11}{*}{\rotatebox[origin=c]{90}{SSIM $\uparrow$}}
& Brain & T2w-to-T2f & BraTS23
    & \cellcolor{hl!60}$\mathbf{0.83}_{0.05}$ & \cellcolor{hl!31}$0.75_{0.05}$ & \cellcolor{hl!33}$\underline{0.76}{}_{0.04}$
    & \cellcolor{hl!3}$0.55_{0.07}$ & \cellcolor{hl!0}$0.52_{0.05}$ & \cellcolor{hl!0}$0.52_{0.06}$ & \cellcolor{hl!7}$0.60_{0.06}$
    & $0.72_{0.06}$ \\
\cmidrule{2-12}
& \multirow{4}{*}{Head-Neck}
  & \multirow{2}{*}{T1w-to-CT} & Synthrad23
    & \cellcolor{hl!60}$\mathbf{0.85}_{0.04}$ & \cellcolor{hl!21}$\underline{0.77}{}_{0.05}$ & \cellcolor{hl!0}$0.66_{0.03}$
    & \cellcolor{hl!21}$\underline{0.77}{}_{0.04}$ & \cellcolor{hl!18}$0.76_{0.04}$ & \cellcolor{hl!21}$\underline{0.77}{}_{0.05}$ & \cellcolor{hl!21}$\underline{0.77}{}_{0.04}$
    & $0.91_{0.04}$ \\
& & & Synthrad25
    & \cellcolor{hl!60}$\mathbf{0.76}_{0.09}$ & \cellcolor{hl!25}$\underline{0.68}{}_{0.07}$ & \cellcolor{hl!6}$0.58_{0.10}$
    & \cellcolor{hl!2}$0.55_{0.05}$ & \cellcolor{hl!1}$0.54_{0.05}$ & \cellcolor{hl!2}$0.55_{0.06}$ & \cellcolor{hl!0}$0.53_{0.06}$
    & $0.69_{0.11}$ \\
\cmidrule(lr){3-12}
& & \multirow{2}{*}{CBCT-to-CT} & Synthrad23
    & \cellcolor{hl!60}$\mathbf{0.89}_{0.04}$ & \cellcolor{hl!44}$\underline{0.87}{}_{0.06}$ & \cellcolor{hl!19}$0.81_{0.07}$
    & \cellcolor{hl!10}$0.77_{0.09}$ & \cellcolor{hl!0}$0.71_{0.11}$ & \cellcolor{hl!1}$0.72_{0.12}$ & \cellcolor{hl!14}$0.79_{0.10}$
    & $0.81_{0.07}$ \\
& & & Synthrad25
    & \cellcolor{hl!60}$\mathbf{0.87}_{0.04}$ & \cellcolor{hl!37}$\underline{0.84}{}_{0.04}$ & \cellcolor{hl!9}$0.75_{0.08}$
    & \cellcolor{hl!2}$0.71_{0.08}$ & \cellcolor{hl!0}$0.70_{0.09}$ & \cellcolor{hl!2}$0.71_{0.09}$ & \cellcolor{hl!3}$0.72_{0.10}$
    & $0.80_{0.09}$ \\
\cmidrule{2-12}
& \multirow{4}{*}{Lung}
  & \multirow{2}{*}{CT-to-PET} & autoPET
    & \cellcolor{hl!60}$\mathbf{0.91}_{0.02}$ & \cellcolor{hl!44}$\underline{0.89}{}_{0.02}$ & \cellcolor{hl!23}$0.84_{0.03}$
    & \cellcolor{hl!0}$0.72_{0.06}$ & \cellcolor{hl!4}$0.75_{0.05}$ & \cellcolor{hl!6}$0.76_{0.06}$ & \cellcolor{hl!4}$0.75_{0.06}$
    & $0.94_{0.03}$ \\
& & & EnhancePET
    & \cellcolor{hl!60}$\mathbf{0.95}_{0.03}$ & \cellcolor{hl!52}$\underline{0.94}{}_{0.04}$ & \cellcolor{hl!45}$0.93_{0.04}$
    & \cellcolor{hl!0}$0.75_{0.08}$ & \cellcolor{hl!0}$0.75_{0.08}$ & \cellcolor{hl!13}$0.83_{0.10}$ & \cellcolor{hl!13}$0.83_{0.10}$
    & $0.90_{0.06}$ \\
\cmidrule(lr){3-12}
& & T1w-to-CT & Synthrad25
    & \cellcolor{hl!60}$\mathbf{0.73}_{0.05}$ & \cellcolor{hl!28}$\underline{0.66}{}_{0.08}$ & \cellcolor{hl!10}$0.57_{0.10}$
    & \cellcolor{hl!13}$0.59_{0.05}$ & \cellcolor{hl!0}$0.49_{0.08}$ & \cellcolor{hl!0}$0.49_{0.09}$ & \cellcolor{hl!10}$0.57_{0.07}$
    & $0.81_{0.06}$ \\
& & CBCT-to-CT & Synthrad25
    & \cellcolor{hl!60}$\mathbf{0.80}_{0.06}$ & \cellcolor{hl!41}$\underline{0.78}{}_{0.06}$ & \cellcolor{hl!8}$0.69_{0.08}$
    & \cellcolor{hl!0}$0.65_{0.10}$ & \cellcolor{hl!0}$0.65_{0.11}$ & \cellcolor{hl!2}$0.66_{0.11}$ & \cellcolor{hl!2}$0.66_{0.11}$
    & $0.74_{0.08}$ \\
\cmidrule{2-12}
& \multirow{2}{*}{Pelvis}
  & T1w-to-CT & Synthrad23
    & \cellcolor{hl!60}$\mathbf{0.88}_{0.02}$ & \cellcolor{hl!21}$\underline{0.80}{}_{0.04}$ & \cellcolor{hl!21}$\underline{0.80}{}_{0.05}$
    & \cellcolor{hl!9}$0.75_{0.05}$ & \cellcolor{hl!0}$0.69_{0.06}$ & \cellcolor{hl!1}$0.70_{0.07}$ & \cellcolor{hl!8}$0.74_{0.05}$
    & $0.87_{0.05}$ \\
& & CBCT-to-CT & Synthrad23
    & \cellcolor{hl!60}$\mathbf{0.88}_{0.03}$ & \cellcolor{hl!34}$\underline{0.85}{}_{0.06}$ & \cellcolor{hl!0}$0.74_{0.04}$
    & \cellcolor{hl!4}$0.76_{0.05}$ & \cellcolor{hl!0}$0.74_{0.07}$ & \cellcolor{hl!0}$0.74_{0.08}$ & \cellcolor{hl!6}$0.77_{0.06}$
    & $0.75_{0.13}$ \\

\bottomrule
\end{tabular*}
 worst~\protect\heatbar~best
\end{table}

\begin{figure}
  \centering
    \includegraphics[width=\textwidth]{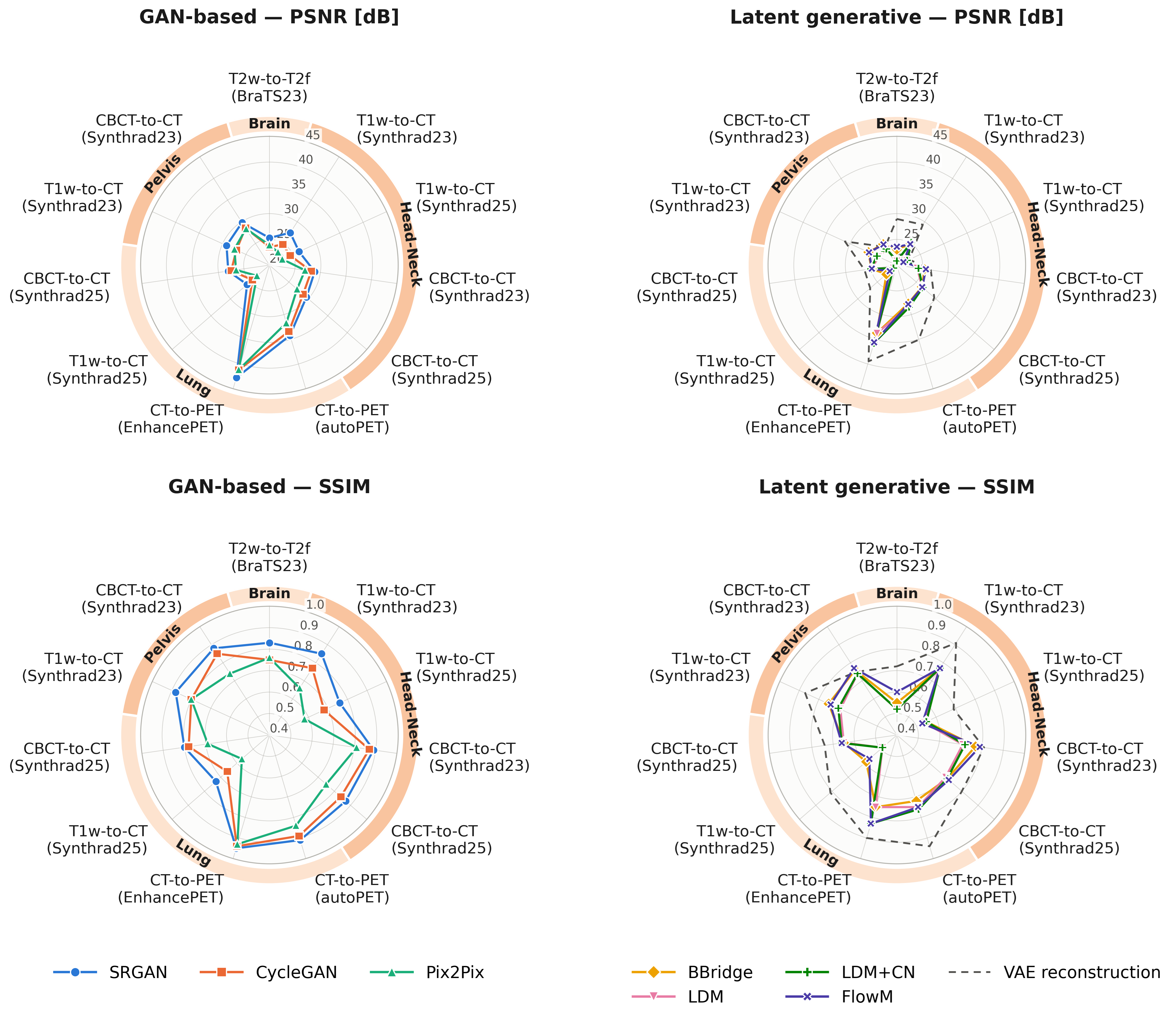}
    \caption{Radar charts of mean PSNR (top) and SSIM (bottom) across the eleven task-anatomy configurations, for GAN-based (left) and latent generative models (right); the dashed grey line is the VAE reconstruction level. Panels of the same metric share the radial scale, which starts at 20 dB for PSNR and at 0.4 for SSIM.}\label{fig3}
\end{figure}

Table~\ref{tbl:generation} reports PSNR and SSIM for the seven architectures across all 11 task–anatomy configurations, with the radar charts in Figure~\ref{fig3} offering a visual summary of the same results.
SRGAN achieves the highest scores on every task.
CycleGAN ranks second on most tasks (8 of 11 for PSNR and 9 of 11 for SSIM), performing well on intra-modality tasks such as CBCT-to-CT, where the domain gap is small because both source and target images share the same imaging modality.
Pix2Pix shows more variable performance across regions, with its accuracy being highly task-dependent. 
On pelvis tasks, it remains competitive, reaching 28.40~dB PSNR on Synthrad23 CBCT-to-CT and 0.80 SSIM on MRI-to-CT, but drops sharply on head/neck tasks, falling to 23.03~dB PSNR and 0.66 SSIM on Synthrad23 MRI-to-CT and to 22.77~dB PSNR and 0.58 SSIM on Synthrad25 MRI-to-CT. 
This inconsistency is especially evident in SSIM, where Pix2Pix ranges from 0.93 on EnhancePET down to 0.57 on Synthrad25 MRI-to-CT (lung), whereas CycleGAN exhibits a narrower range across the same tasks (0.94 to 0.66).
Latent generative models underperform their GAN counterparts: SRGAN exceeds the best latent model by 1.73 to 7.00~dB in PSNR and by 0.08 to 0.24 in SSIM, depending on the task.
The gap is pronounced on structurally complex tasks such as BraTS23 T2w-to-T2f, where latent models reach an SSIM of 0.52--0.60 against 0.83 for SRGAN.
Within the latent family, performance varies across tasks and metrics: LDM+ControlNet achieves the highest latent PSNR on 6 of the 11 tasks, whereas FlowM achieves the highest latent SSIM on 7.
The addition of ControlNet to the LDM baseline yields PSNR gains of at most 0.37~dB on 10 of the 11 tasks (e.g., 24.25 vs.\ 23.98~dB on Synthrad23 CBCT-to-CT, head/neck), and SSIM gains that are consistent across tasks but small (at most 0.015), suggesting that conditioning alone does not compensate for the information loss following latent-space compression.
The only exception is the EnhancePET CT-to-PET task, where LDM+ControlNet reaches 35.77~dB and 0.83 SSIM compared to 33.79~dB and 0.75 for the plain LDM.

A key factor contributing to the consistently lower performance of latent models is the loss of reconstruction fidelity introduced by the VAE's encoding–decoding stage.
As reported in Table~\ref{tbl:generation} (last column), the VAE reconstruction alone introduces measurable fidelity loss across all datasets. 
For instance, on Synthrad23 MRI-to-CT (head/neck), the VAE achieves a PSNR of 29.33~dB and an SSIM of 0.91, values that represent a reference no latent generative model can on average exceed. 
The effect is particularly severe on datasets with high anatomical variability: on Synthrad25 MRI-to-CT (head/neck), the VAE reconstruction reaches only 22.66~dB PSNR and 0.69 SSIM, leaving little room for the downstream latent generative model to operate without further degrading quality. 
This confirms that the autoencoder backbone is a bottleneck for tasks requiring accuracy in pixel space~\cite{rombach2022}. 
We notice that, for some datasets, the VAE reconstruction is lower than the translation achieved by certain latent models (e.g., Synthrad23 CBCT-to-CT, pelvis: VAE has a PSNR of 24.22 dB vs. BBridge, 24.84 dB).
The VAE reconstruction score measures how faithfully the target image is recovered after encoding and decoding. It provides a reference for autoencoder fidelity, but not a strict upper bound on translation performance: a generative model may predict a latent representation that decodes more accurately than the target image's own encoding. 
When source and target are structurally similar (as in CBCT-to-CT), the latent model can learn a near-identity mapping that partially compensates for VAE reconstruction artifacts.
This could explain why the VAE reference reconstruction level represents an upper bound on average but may be exceeded for individual tasks where the source provides strong structural guidance.

The lower fidelity of latent models comes at a higher inference cost (see Supplementary Materials, Section~\ref{app:performance}). 
Median inference times across tasks range from 22.9 to 132.5~s per volume for latent models, compared with 2.56~s for SRGAN, corresponding to approximately 9--52-fold slower inference. 
The difference reflects both the larger networks used by latent generative models and their iterative sampling for every sliding-window patch. 
Under the evaluated configurations, SRGAN therefore combines higher image fidelity with faster inference.

\begin{table}[h]
\caption{Pairwise Wilcoxon signed-rank test on model rankings (PSNR and SSIM). Each cell reports the number of tasks (out of 11) in which the model on the row ranked higher than the model on the column. Asterisks (*) denote significant differences after Holm--Bonferroni correction (two-sided exact test, $\alpha=0.05$).}\label{tbl5}
\fontsize{8}{9.5}\selectfont
\begin{tabular*}{\tblwidth}{@{\extracolsep{\fill}}lccccccc}
\toprule
\multicolumn{8}{c}{\textbf{(a) PSNR}} \\ \midrule
vs. & SRGAN & CycleGAN & Pix2Pix & BBridge & LDM & LDM+CN & FlowM \\ \midrule
SRGAN & --- & 11* & 11* & 11* & 11* & 11* & 11* \\
CycleGAN & 0 & --- & 9 & 10 & 10* & 10 & 9 \\
Pix2Pix & 0 & 2 & --- & 10 & 10 & 10 & 10 \\
BBridge & 0 & 1 & 1 & --- & 6 & 5 & 6 \\
LDM & 0 & 1 & 1 & 5 & --- & 3 & 4 \\
LDM+CN & 0 & 1 & 1 & 6 & 8 & --- & 6 \\
FlowM & 0 & 2 & 1 & 5 & 7 & 5 & --- \\
\midrule
\multicolumn{8}{c}{\textbf{(b) SSIM}} \\ \midrule
vs. & SRGAN & CycleGAN & Pix2Pix & BBridge & LDM & LDM+CN & FlowM \\ \midrule
SRGAN & --- & 11* & 11* & 11* & 11* & 11* & 11* \\
CycleGAN & 0 & --- & 9 & 10* & 11* & 10* & 10 \\
Pix2Pix & 0 & 1 & --- & 8 & 10* & 10 & 9 \\
BBridge & 0 & 1 & 3 & --- & 8 & 5 & 3 \\
LDM & 0 & 0 & 1 & 3 & --- & 0 & 2 \\
LDM+CN & 0 & 1 & 1 & 6 & 11* & --- & 2 \\
FlowM & 0 & 1 & 2 & 8 & 9* & 9 & --- \\
\bottomrule
\end{tabular*}
\end{table}

The dominance of GANs is modulated by task difficulty. 
Intra-modality CBCT-to-CT synthesis outperforms its MRI-to-CT counterpart within matched anatomical regions and datasets, suggesting that a structural proximity between source and target domains facilitates generation. 
MRI-to-CT translation proves more challenging, with the performance gap between GAN-based and latent architectures widening as anatomical complexity increases. 
The divergence between the two families emerges on the T2w-to-T2f task: while GAN-based models maintain high structural similarity, all latent models drop, a result that aligns with the VAE reconstruction level discussed above, as the subtle contrast differences between T2-w and T2-f sequences could be vulnerable to information loss in the encoding–decoding stage.
CT-to-PET synthesis requires a separate consideration. 
Although it yields the highest absolute PSNR and SSIM values across all experiments, these scores are misleading: the inherently sparse signal of PET images, with large background regions, artificially inflates both pixel-wise and structural fidelity metrics independently of the actual synthesis quality. 
This is a well-known limitation of standard image quality metrics~\cite{roberts2023, tang2024}, particularly pronounced for functional modalities such as PET, where clinically meaningful information is concentrated in a small fraction of voxels.

The pairwise Wilcoxon signed-rank analysis in Table~\ref{tbl5} provides statistical grounding for these observations. 
After Holm correction, GAN-based approaches dominate the ranking across both PSNR and SSIM, with SRGAN achieving the best task-level ranking across all 11 configurations.
Within the latent family, pairwise differences are rarely statistically significant, further confirming the absence of a clear winner among these approaches. 
The result indicates that the hierarchical, multi-scale structure of medical volumetric data is an important property to account for when designing generative architectures.
This finding generalizes the observations of Ha et al. \cite{ha2025}, who demonstrated the advantage of multi-resolution processing on a smaller set of tasks.

\subsection{Tumor-Level Fidelity Evaluation}\label{sec:tumor_level}

Beyond the whole-volume evaluation presented in the previous section, we assess how faithfully the models preserve tumor features, as synthesis artifacts within these regions could compromise the diagnostic reliability of the generated images. 
To this end, we use the ground-truth segmentation masks available for BraTS23 (T2w-to-T2f) and autoPET (CT-to-PET) datasets to compute metrics in tumor regions.
It is worth noting that such two datasets differ in their lesion characteristics: BraTS23 contains relatively large brain tumors (median diameter 51.2~mm, interquartile range, IQR, 37.9–62.3~mm), while autoPET is dominated by smaller pulmonary lesions (median diameter 19.3~mm, IQR 15.2–30.3~mm), making the latter a more demanding challenge for tumor-level synthesis.

\begin{figure}
  \centering
    \includegraphics[width=0.9\textwidth]{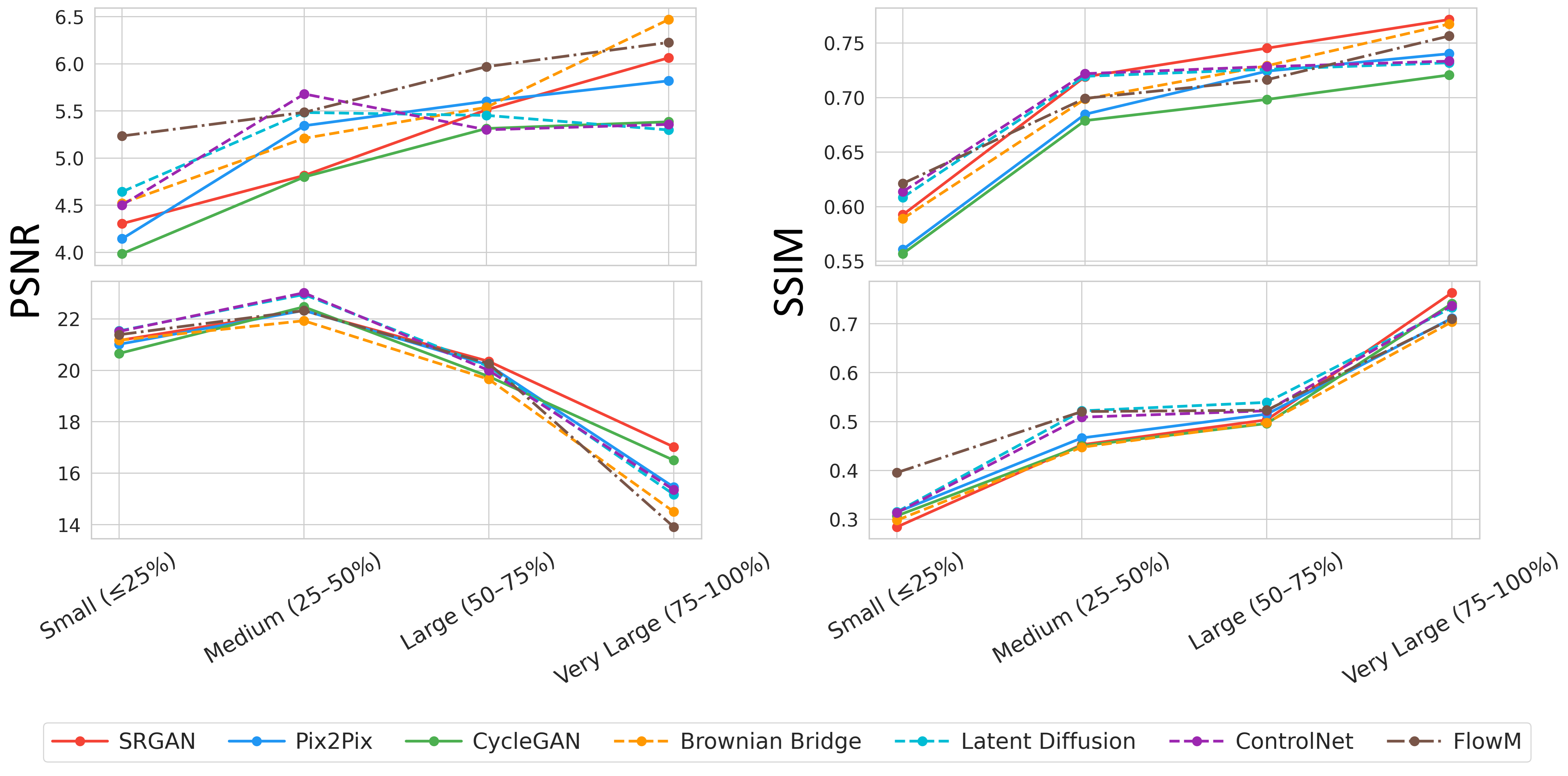}
    \caption{Lesion analysis from BraTS23 (MRI T2w-to-T2f, top) and autoPET (CT-to-PET, bottom). PSNR (left) and SSIM (right) vs lesion size group (for BraTS, median lesion diameter: 51.2 mm, IQR: 37.9–62.3 mm; for autoPET, median lesion diameter: 19.3 mm, IQR: 15.2–30.3 mm).} \label{fig:lesions}
\end{figure}

Figure~\ref{fig:lesions} reports PSNR (left) and SSIM (right) as a function of lesion size for the two datasets, respectively. 
For the BraTS23 dataset (Figure~\ref{fig:lesions}, top), both metrics improve with increasing lesion size across all models, confirming that larger lesions are easier to synthesize due to their greater spatial extent and this benefit is uniform across both metrics. 
For the autoPET dataset (Figure~\ref{fig:lesions}, bottom), a different pattern emerges: indeed, while SSIM improves with lesion size, suggesting that models successfully capture the morphological structure of larger lesions, PSNR decreases. 
The divergence between structural and pixel-level metrics suggests that the model preserves lesion shape more accurately than absolute intensity values, highlighting that structural fidelity does not necessarily imply accurate uptake quantification in CT-to-PET translation. 
The smaller lesions in this dataset, with a median diameter approximately one-third of that in BraTS23, further complicate the task by providing fewer lesion voxels from which to infer the metabolic signal.
These results highlight two complementary challenges in tumor-level synthesis. 
Smaller lesions are more difficult to reproduce across models, potentially because their limited spatial extent provides fewer informative voxels for learning. 
In CT-to-PET translation, the additional challenge of inferring metabolic activity from anatomical information leads to unreliable intensity predictions even when lesion structure appears well preserved (as SSIM scores suggest). 
Together, these findings show that preserving lesion morphology alone is insufficient to ensure faithful reproduction of its functional signal.
Given the observed degradation in intensity accuracy, synthetic PET generated by the evaluated models should not currently be relied upon for quantitative SUV-based clinical decisions without dedicated validation of uptake accuracy.

\subsection{Perceptual Evaluation}\label{sec:perceptual}

We now turn to the analysis of perceptual realism, comparing quantitative rankings with human perception collected via the Visual Turing test administered to 17 physicians. 
This analysis probes two questions: whether the models that score highest on PSNR and SSIM also produce the volumes that clinicians consider most realistic, and whether expert readers can distinguish synthetic acquisitions from real ones. 
To this aim, we use the synthetic volumes generated by the two best-performing GAN-based models (SRGAN and CycleGAN) and the two best-performing latent generative models (BBridge and FlowM). 
Table~\ref{tab:vtt_stats} summarizes per-task classification accuracy with 95\% confidence intervals and inter-reader agreement for all three parts of the test.

\begin{table}[t]
\centering
\caption{Visual Turing test statistics. \textbf{(a)} Part~1 classification accuracy per translation task, reported separately for real and AI-generated volumes and overall, with 95\% confidence intervals obtained by reader-level bootstrap (5000 resamples). Inter-reader agreement is quantified with Fleiss' $\kappa$. \textbf{(b)} Inter-reader agreement for Part~2 (Fleiss' $\kappa$ on the three response categories) and Part~3 (Kendall's $W$ averaged across triplets), with 95\% confidence intervals obtained by bootstrap over items (5000 resamples).}
\label{tab:vtt_stats}
\small
\begin{tabular}{lccccc}
\toprule
\multicolumn{6}{l}{\textit{(a) Part 1: real vs.\ AI-generated classification}} \\
\midrule
Task & Items (R/S) & Acc.\ real (\%) & Acc.\ synthetic (\%) & Acc.\ overall (\%) [95\% CI] & Fleiss' $\kappa$ \\
\midrule
T1w-to-CT   & 1/3 & 62.5 & 60.4 & 60.9\ [48.4, 75.0] & 0.01 \\
CBCT-to-CT  & 2/1 & 68.8 & 87.5 & 75.0\ [66.7, 83.3] & 0.33 \\
T2w-to-T2f  & 2/2 & 46.9 & 43.8 & 45.3\ [37.5, 53.1] & 0.05 \\
CT-to-PET   & 2/2 & 53.1 & 46.9 & 50.0\ [37.5, 62.5] & $-0.03$ \\
\midrule
All tasks   & 7/8 & 57.1 & 56.2 & 56.7\ [51.7, 62.1] & 0.08 \\
\midrule
\multicolumn{6}{l}{\textit{(b) Parts 2 and 3: inter-reader agreement}} \\
\midrule
\multicolumn{4}{l}{Part 2, GAN pairs (SRGAN vs.\ CycleGAN), Fleiss' $\kappa$} & \multicolumn{2}{c}{0.15\ [0.03, 0.25]} \\
\multicolumn{4}{l}{Part 2, latent pairs (FlowM vs.\ BBridge), Fleiss' $\kappa$} & \multicolumn{2}{c}{0.15\ [$-0.02$, 0.29]} \\
\multicolumn{4}{l}{Part 3, ranking triplets, Kendall's $W$}                   & \multicolumn{2}{c}{0.27\ [0.18, 0.37]} \\
\bottomrule
\end{tabular}
\end{table}

In the first part of the test, participants classified each volume as real or AI-generated. 
Overall accuracy was 57\% (95\% CI 52--62), above the chance level of 50\%, indicating that readers could marginally distinguish synthetic volumes from real acquisitions. 
Misclassification errors were balanced in both directions, with 43\% of real images incorrectly labeled as AI-generated and 44\% of synthetic images mistaken for real ones. 
Accuracy varied with the displayed modality: MRI images were the most challenging to classify correctly (45\%), followed by PET (50\%) and CT (67\%). 
At the task level (Table~\ref{tab:vtt_stats}a), accuracy was highest for CBCT-to-CT (75.0\%, 95\% CI 66.7--83.3), where 87.5\% of readers recognized the synthetic volume, whereas it did not differ from chance for T1w-to-CT (60.9\%, 48.4--75.0) and CT-to-PET (50.0\%, 37.5--62.5) and was lowest for T2w-to-T2f (45.3\%, 37.5--53.1). 
Inter-reader agreement was slight (Fleiss' $\kappa = 0.08$), showing that readers did not converge on a shared set of cues for identifying synthetic volumes.

\begin{figure}
  \centering
    \includegraphics[width=0.7\textwidth]{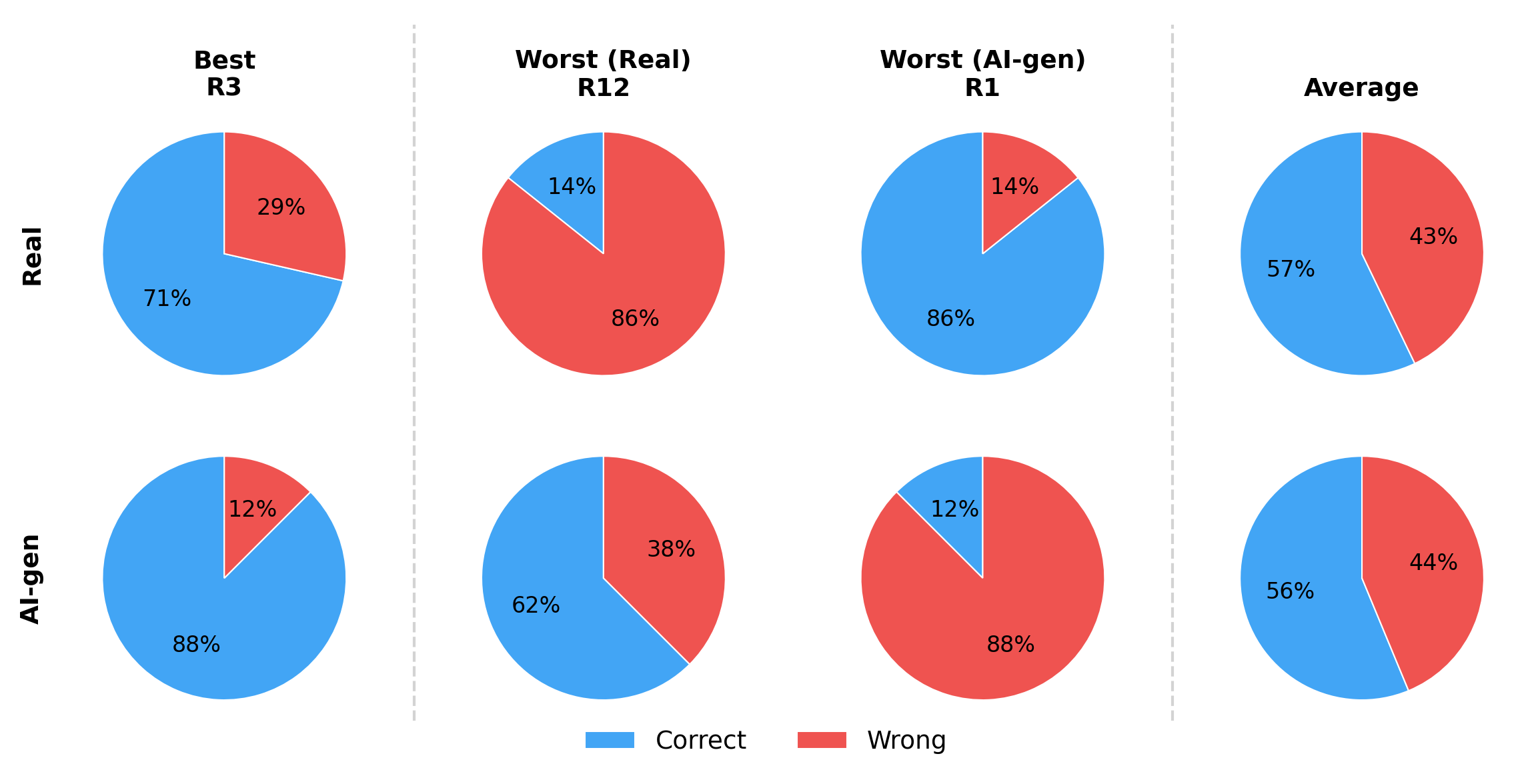}
    \caption{ Summary of classification performance in Visual Turing test, Part~1. 
Each column reports the rate of correctly (blue) and incorrectly 
(red) classified images from one physician, separately for real and AI-generated cases (top and bottom row, respectively). \textbf{Best}: physician with the highest balanced 
accuracy (R3). \textbf{Worst (Real)} and \textbf{Worst (AI-gen)}: 
physicians with the lowest accuracy on real and AI-generated images, 
respectively (R12 and R1), illustrating opposite classification biases. 
\textbf{Average}: mean performance across all physicians.} \label{fig:vtt_part1}
\end{figure}

Individual performance is summarized in Figure~\ref{fig:vtt_part1}. 
Seven of the 17 readers exceeded 50\% accuracy on both real and AI-generated images.
Physician R3 achieved the highest accuracy (80\%; 71\% on real and 88\% on AI-generated images) and was the only reader above chance at the nominal level (binomial $p=0.035$), although not after Holm correction for the 17 readers. 
No single physician performed below 50\% on both classes; the two worst cases instead reflected opposite classification biases. 
Physician R12 correctly identified only 14\% of real images, systematically labeling them as AI-generated, whereas R1 correctly identified only 12\% of AI-generated images, systematically labeling them as real. 
Together, these two cases show that low accuracy on one class can coexist with above-average performance on the other, revealing two distinct failure modes within the cohort. 
Neither years of experience nor subspecialty had a measurable impact on classification accuracy (more than 10 years of practice: 58.8\% vs.\ 52.0\%, Mann--Whitney $p=0.30$; experience with PET: 56.7\% vs.\ 56.7\%, $p=0.87$), although the small subgroups limit this comparison.

\begin{figure}
  \centering
    \includegraphics[width=0.9\textwidth]{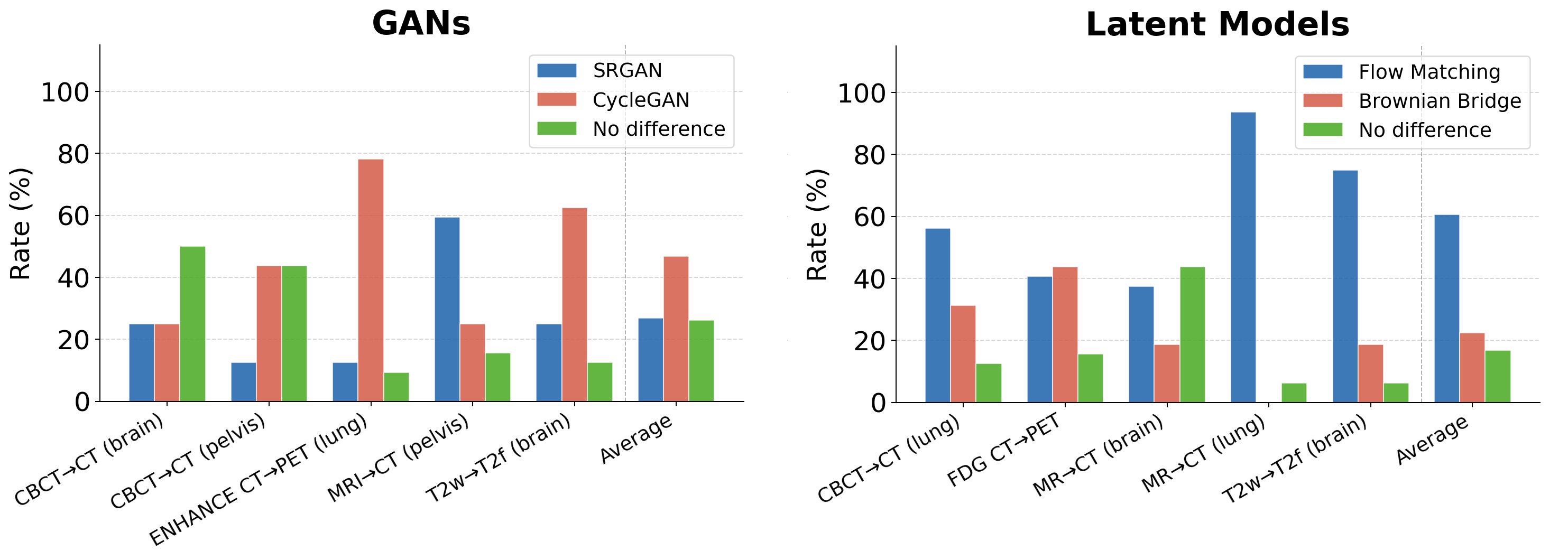}
    \caption{Pairwise preference results (Visual Turing test, Part~2). Results are reported for GANs (CycleGAN and SRGAN, left) and latent models (Brownian Bridge and Flow Matching, right), for each task and as an overall aggregate ("Average").} \label{fig:vtt_part2}
\end{figure}

The second part of the test asked participants to compare pairs of synthetic volumes and select the one they preferred, or indicate that no difference could be identified. 
Figure~\ref{fig:vtt_part2} reports the results of direct comparisons within the GAN family (SRGAN vs.\ CycleGAN) and within the latent family (BBridge vs.\ FlowM); preferences were tested per reader with the Wilcoxon signed-rank test, with Holm correction across all Part~2 comparisons. 
Within the GAN family, CycleGAN was chosen more often than SRGAN (46.9\% vs.\ 26.9\%, with 26.2\% reporting no difference), despite SRGAN achieving higher PSNR and SSIM scores, although this overall preference did not reach significance (adjusted $p=0.42$, $r=0.57$). 
The preference is not uniform across tasks and is significant only in CT-to-PET synthesis, where CycleGAN was chosen by 78.1\% of readers against 12.5\% for SRGAN (adjusted $p=0.011$, $r=0.96$), suggesting that in this task CycleGAN produces outputs that, while numerically less accurate, appear more natural to the clinical eye. 
In the other tasks the differences were not significant: CycleGAN was favored in T2w-to-T2f (62.5\% vs.\ 25.0\%) and in CBCT-to-CT, pelvis (43.8\% vs.\ 12.5\%, with an equally high proportion of ``no difference'' responses, 43.8\%), whereas SRGAN was favored in MRI-to-CT (pelvis) (59.4\% vs.\ 25.0\%). 
CBCT-to-CT (brain) constitutes a borderline case, with 50.0\% of readers reporting no perceptual difference and the remaining preferences split evenly between the two models (25.0\% each).
Among latent models, FlowM was preferred over BBridge (60.6\% vs.\ 22.5\%, with 16.9\% reporting no difference; adjusted $p=0.030$, $r=0.79$), consistent with its higher SSIM on 8 of the 11 tasks. FlowM is most strongly favored in MRI-to-CT (lung), where it was selected by 93.8\% of readers, with no preference for BBridge and only 6.2\% reporting no difference (adjusted $p<0.001$, $r=1.00$). 
It also leads, without reaching significance after correction, in T2w-to-T2f (brain) (75.0\% vs.\ 18.8\%, 6.2\% no difference) and in CBCT-to-CT (lung) (56.2\% vs.\ 31.2\%, 12.5\% no difference). 
In MRI-to-CT (brain), FlowM retains a nominal lead (37.5\% vs.\ 18.8\%), but the largest share of responses falls in the ``no difference'' category (43.8\%). CT-to-PET is the only case where BBridge is marginally preferred (43.8\% vs.\ 40.6\%, 15.6\% no difference), making the two models nearly indistinguishable to expert readers. 
Inter-reader agreement on pairwise preferences was slight for both GAN pairs ($\kappa = 0.15$) and latent pairs ($\kappa = 0.15$). 
On the three control questions of identical images, 75\% of responses correctly reported no perceptual difference, while 25\% indicated a preference; these responses came from half of the readers (8 of 17), suggesting a degree of response bias that should be considered when interpreting small preference differences.

These pairwise comparisons show that the model achieving the highest quantitative scores is not necessarily the one preferred by clinicians, most clearly in CT-to-PET synthesis, reinforcing the need for perceptual evaluation as a complement to standard metrics. 
The agreement between metrics and clinical preference also depends on the nature of the translation task. 
In CBCT-to-CT synthesis, an intra-modality task with a narrow domain gap, clinicians frequently perceive no meaningful difference between competing models (50.0\% ``no difference'' in the brain and 43.8\% in the pelvis for the GAN comparison), suggesting that model selection is not critical when outputs reach similar acceptable quality. 
This perception is consistent with the quantitative evaluation (Table~\ref{tbl:generation}), where CBCT-to-CT shows the smallest differences between SRGAN and CycleGAN (0.6--1.2~dB in PSNR, against 1.4--2.7~dB for T1w-to-CT). In MRI-to-CT translation, the inter-modality gap amplifies differences between architectures. 
SRGAN is preferred by 59.4\% of readers in the pelvis, a trend in line with its higher PSNR and SSIM scores that may reflect a more accurate reproduction of bone and soft-tissue contrast, although it did not reach significance. 
Among latent models, FlowM dominates in MRI-to-CT (lung) with 93.8\% preference, whereas in MRI-to-CT (brain) 43.8\% of readers report no difference, suggesting that the perceptual gap may narrow in regions with less anatomical complexity.

CT-to-PET synthesis is the most revealing case. 
CycleGAN is strongly preferred over SRGAN (78.1\% vs.\ 12.5\%), while FlowM and BBridge receive similar perceptual scores (40.6\% vs.\ 43.8\%) despite FlowM's higher SSIM on this task. 
This suggests that clinicians evaluate PET image quality primarily on structural plausibility rather than pixel-wise intensity accuracy, judging whether lesion morphology and anatomical context appear realistic while being less sensitive to deviations in absolute SUV values.

\begin{figure}
  \centering
    \includegraphics[width=0.9\textwidth]{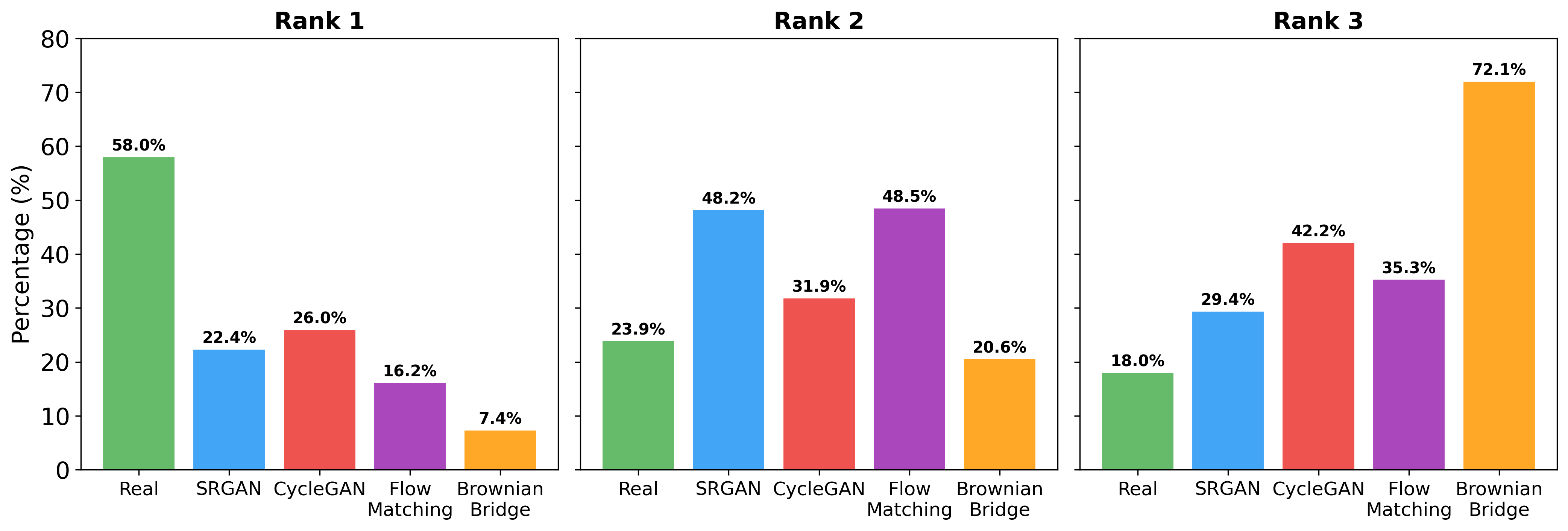}
    \caption{Three-way ranking results (Visual Turing test, Part~3). Each panel reports the percentage of rank 1 (most realistic), rank 2, and rank 3 (least realistic) assignments across triplets, aggregated over tasks and physicians.} \label{fig:vtt_part3}
\end{figure}

The third part of the test asked participants to rank triplets of volumes from most to least realistic (rank 1 to 3), with results aggregated over all tasks and physicians in Figure~\ref{fig:vtt_part3}. 
Real volumes were ranked as most realistic in 58\% of cases (95\% CI 52--64), well above the one-in-three expected by chance; conversely, in 42\% of the triplets a synthetic image was judged more realistic than the corresponding real acquisition. 
In direct comparisons within the same triplet, the real image was ranked above SRGAN (68\% of comparisons), CycleGAN (67\%) and FlowM (71\%) significantly more often than the reverse (adjusted $p\le0.041$). Among synthetic models, CycleGAN obtained the highest frequency of rank-1 assignments (26.0\%) and SRGAN the second highest (22.4\%), but CycleGAN was also ranked last more often (42.2\% vs.\ 29.4\%), and when both appeared in the same triplet SRGAN was ranked above CycleGAN in 56\% of comparisons, a difference that was not significant ($p=0.33$); the ranking task therefore does not show a perceptual preference between the two GAN models. 
SRGAN and FlowM were the most frequently second-ranked models, each assigned rank 2 in approximately half of the comparisons, whereas BBridge received rank 3 in 72\% of cases and was ranked below FlowM and the real image in all or nearly all direct comparisons, confirming its weakest perceptual quality among the evaluated architectures. 
Concordance among readers' rankings was weak (Kendall's $W = 0.27$). 
The two control questions, each containing one real image and two outputs from the same model, exposed a modality-dependent difference in discriminability. 
In the CT-to-PET triplet, 70.6\% of readers ranked an SRGAN-generated image as the most realistic, failing to identify the real acquisition, in line with the difficulty of discriminating synthetic PET images observed throughout the test. 
In the T2w-to-T2f triplet, only 17.6\% of readers were fooled, suggesting that MRI synthesis artifacts remain more perceptible to expert readers.

Taken together, the three parts of the Visual Turing test show that the models with the highest scores on standard metrics do not necessarily produce the images that physicians prefer, as observed most clearly for CycleGAN and SRGAN in CT-to-PET synthesis. 
When judging single volumes, readers were only slightly better than chance at identifying synthetic images, and in 42\% of the triplets a synthetic image was ranked above the real acquisition, indicating that current generative models, particularly GAN-based architectures, have reached a level of visual fidelity that often challenges expert discrimination. 
When real and synthetic volumes are compared side by side, however, readers still recognize the real acquisition more often than chance, showing that residual differences remain perceptible.

\section{Limitations and Future Directions}\label{sec:limitations}
While this work provides a comprehensive comparison of generative architectures for volumetric medical image translation, it is important to acknowledge limitations to identify directions for future research.

First, the scarcity of public cancer datasets with paired multi-modal volumes~\cite{friedrich2024} limits the translation directions that could be evaluated (Figure~\ref{fig1}).
Annotated datasets are even scarcer: of the 11 configurations included, only two have lesion segmentation masks, restricting the tumor-level analysis to a subset of tasks and preventing a systematic assessment of pathological fidelity across all translation directions.
Expanding the availability of public multi-modal datasets with voxel-level annotations remains a prerequisite for moving the field beyond proof-of-concept evaluations.

Second, the architectural comparison is affected by the shared frozen VAE, which compresses information and introduces reconstruction losses absent in GANs, which operate directly in image space. 
Future work should investigate alternative autoencoders, compression factors, and multi-resolution latent representations to better preserve anatomical detail.
Indeed, quantitative results indicate that a multi-resolution framework is better suited to the multi-scale structure of the input data for medical image translation. 
SRGAN~\cite{ha2025} achieves the best performance thanks to its multi-resolution framework, which processes data at multiple spatial scales and captures anatomical information that single-scale architectures miss. 
Extending this principle to latent generative models, for instance through multi-scale VAE encoders capable of preserving features across compression levels, or through hierarchical latent spaces that retain both global anatomy and local detail, represents a promising direction for future research.

Third, inter-modality tasks such as CT-to-PET, where source and target encode fundamentally different physical quantities, expose the limits of current approaches. 
Models can reproduce lesion morphology but are less reliable in predicting metabolic intensity, as SUV values depend on physiological information that CT attenuation maps do not carry, and addressing tasks of this complexity will likely require domain-specific innovations such as physics-informed losses or hybrid architectures that incorporate auxiliary physiological priors. 
Given the observed degradation in intensity accuracy, synthetic PET generated by the evaluated models should not currently be relied upon for quantitative SUV-based clinical decisions without dedicated validation of uptake accuracy.

Fourth, the evaluation framework itself presents open challenges. 
As demonstrated by the Visual Turing test, standard metrics such as PSNR and SSIM do not reliably reflect human perception~\cite{dayarathna2024, friedrich2024}, a finding that is supported by recent literature~\cite{breger2025, dohmen2025}, highlighting that these metrics were designed for computer vision tasks and exhibit shortcomings when applied to medical data, including insensitivity to blurring, dependence on normalization methods, and poor correlation with human judgment. 
Developing new evaluation metrics specifically designed for 3D medical volumes, capable of capturing local anatomical accuracy, functional signal fidelity, and sensitivity to pathological features, remains a critical open challenge in the field~\cite{friedrich2024}.

Finally, the Visual Turing test used viewing conditions that differ from routine clinical practice, highlighting the need for more clinically representative viewing tools in future reader studies.

\section{Conclusion}\label{sec:conclusion}
Virtual scanning has the potential to complement routine oncological imaging by synthesizing information from modalities not acquired during the examination. 
This could reduce the need for further exams and improve access to advanced imaging. 
However, realizing this requires evidence that synthetic images preserve clinically relevant information. 
Within this context, this work presents a systematic benchmark of seven generative architectures for 3D medical I2I translation, evaluated across 77 experiments spanning five oncological datasets, four translation directions, and three anatomical regions (head/neck, lung, and pelvis) covering a broad range of cancer types. 

Under the evaluated configurations, SRGAN achieves the highest PSNR and SSIM across all tasks. 
Latent generative models achieve lower quantitative fidelity, with reconstruction losses introduced by the frozen VAE contributing to the performance gap.
These findings suggest that multi-resolution processing may benefit volumetric synthesis, motivating further investigation of multi-scale architectures and alternative latent representations.

The tumor-level analysis shows that fidelity depends on lesion size and translation task. Both PSNR and SSIM improve with increasing lesion size in BraTS23, whereas CT-to-PET synthesis exhibits increasing SSIM alongside decreasing PSNR. This divergence suggests that improved structural similarity does not necessarily imply accurate reproduction of lesion intensities, highlighting the need for direct assessment of metabolic uptake fidelity.

The Visual Turing test shows that quantitative rankings do not align with expert preferences. CycleGAN is preferred more frequently than SRGAN overall despite its lower quantitative scores, although the difference is statistically significant only for CT-to-PET. These findings support perceptual assessment as a complement to standard metrics, while the limited reader sample and differences from routine clinical viewing conditions constrain their interpretation.

This study provides a reusable evaluation protocol to support future model development, dataset expansion, and clinically grounded validation in medical image synthesis. The full codebase, including model implementations, training pipelines, evaluation scripts, and the Visual Turing test platform, will be published upon acceptance.

\section*{Acknowledgments}
This work was supported by Strategic University Projects “DIGILUNG: AI-boosted DIGital twIn for personaLized lUNG cancer care, Kempe Foundation
under grant no. JCSMK24-0094; and "AI-powered
Digital Twin for next-generation lung cancEr cAre” financed by Universit`a Campus
Bio- Medico di Roma 2023 (GEN0469). 
Resources are provided by the National
Academic Infrastructure for Supercomputing in Sweden (NAISS) and the Swedish
National Infrastructure for Computing (SNIC) at Alvis @ C3SE, partially funded by
the Swedish Research Council through grant agreements no. 2022-06725 and no. 2018-
05973. 
The funders had no role in study design, data collection and analysis, decision to publish, or preparation of the manuscript.

\printcredits

\bibliographystyle{model1-num-names}

\bibliography{cas-refs}

\newpage

\appendix

\section{Dataset pre-processing}\label{app:datasets}

Pre-processing starts from the procedures provided by the dataset curators, followed by a custom pipeline developed for this benchmark and implemented in PyTorch using MONAI~\cite{monai}.
The resulting source–target volumes are used to extract paired 3D patches for model training.

\subsection{Pre-processing provided by dataset curators}

For SynthRAD23 and SynthRAD25, the release procedures include DICOM-to-NIfTI conversion, resampling, rigid inter-modality registration using Elastix~\cite{klein2010elastix}, and anonymisation, including defacing of brain images. 
The reported voxel spacings are $1{\times}1{\times}1$~mm$^3$ for brain images and $1{\times}1{\times}2.5$~mm$^3$ for pelvic images.
Each image pair is accompanied by a binary body mask generated using thresholding and hole filling with the ITK toolkit, then dilated to retain a surrounding air margin.

For BraTS23, all four MRI sequences undergo DICOM-to-NIfTI conversion, co-registration to the SRI24 multi-channel atlas of the normal adult human brain~\cite{rohlfing2010}, resampling to an isotropic resolution of $1{\times}1{\times}1$~mm$^3$, and anonymisation. 
Atlas registration provides a common spatial reference across modalities and subjects.

For AutoPET, pre-processing includes conversion to NIfTI format and resampling of CT volumes to the PET voxel grid. PET intensities are converted to Standardized Uptake Values (SUV)~\cite{kinahan2010positron}. PET volumes are subsequently clipped to $[0,20]$~SUV to retain the intensity range relevant to the analysis. CT volumes are retained without windowing at this stage and undergo intensity clipping in the common pipeline described below.

For EnhancePET, raw PET intensities are provided in $\mathrm{Bq/mL}$. Patient body weight and injected dose are retrieved from the accompanying metadata spreadsheet and used to compute SUV values as follows:
\begin{equation}
\label{eq}
\mathrm{SUV} = \mathrm{PET}_{\mathrm{Bq/mL}}
\times \frac{W \cdot 1000}{D \cdot 10^{6}},
\end{equation}
where $W$ is body weight in kg and $D$ is the injected dose in MBq.

Because CT and PET voxel spacings vary across scanners and centres, CT volumes are resampled to the corresponding PET grid using linear interpolation. Regions outside the CT field of view are filled with $-1024$~HU, corresponding to air. This step aligns both modalities on a common spatial grid. The provided lung masks are resampled to the same grid and used to restrict both volumes to the thoracic region. Specifically, the first and last axial slices containing lung voxels define the crop boundaries, which are extended by 20~mm in both directions to retain surrounding anatomy. The same axial crop is applied to CT and PET, while preserving the full transverse field of view.

\subsection{Pre-processing developed for this benchmark}

\begin{table}
\caption{Voxel spacing (mm) after resampling and foreground intensity thresholds 
(applied after normalization) used to compute the foreground masks for source 
and target volumes.}\label{tblA}
\begin{tabular}{cccc}
\toprule
\textbf{Dataset} & \textbf{Voxel spacing (x$\times$y$\times$z)} & \textbf{Source threshold} & \textbf{Target threshold} \\
\midrule
Synthrad23 (Head/Neck)   & $1.0 \times 1.0 \times 1.0$ & 0.1 & 0.1 \\
Synthrad23 (Pelvis)  & $1.0 \times 1.0 \times 2.5$ & 0.1 & 0.1 \\
Synthrad25 (Lung)    & $1.0 \times 1.0 \times 3.0$ & 0.1 & 0.1 \\
Synthrad25 (Head/Neck) & $1.0 \times 1.0 \times 1.0$ & 0.1 & 0.1 \\
BraTS23            & $1.0 \times 1.0 \times 1.0$ & 0.1 & 0.1 \\
autoPET              & $2.0 \times 2.0 \times 3.0$ & 0.1 & 0.1 \\
EnhancePET     & $4.063 \times 4.063 \times 3.0$ & 0.1 & 0.01 \\
\bottomrule
\end{tabular}
\end{table}

The pipeline comprises the following operations, illustrated in the pre-processing module of Figure~\ref{fig3}. First, body masks, when available, are applied independently to the source and target volumes, and background voxels are replaced with modality-specific values. For CT, CBCT, and PET, the fill value is the minimum intensity within the foreground; for MRI, it is zero.
All volumes are then resampled to the dataset-specific voxel spacings reported in Table~\ref{tblA} using trilinear interpolation. CT and CBCT intensities are clipped to $[-1024,3000]$~HU, and FDG-PET intensities to $[0,20]$~SUV. No fixed clipping range is applied to MRI because its absolute intensities depend on the acquisition protocol. Intensities are subsequently standardised to zero mean and unit variance, and spatial padding is applied to accommodate patch extraction.
Foreground masks are computed independently for the source and target volumes by thresholding the normalised intensities at $0.1$. For EnhancePET, the target threshold is reduced to $0.01$ because a threshold of $0.1$ excludes a substantial fraction of the PET foreground owing to its more compressed intensity distribution after normalisation. The source and target masks are intersected to define a shared foreground region.
Finally, three paired patches of size $96{\times}96{\times}96$ voxels are randomly extracted from each training volume. Patch centres are sampled within the shared foreground mask to avoid patches dominated by background, and identical spatial coordinates are used for the source and target patches.

\section{Implementation Details}\label{app:implementation}

This section provides full architectural details for each of the seven benchmarked models and the shared VAE, complementing the high-level descriptions and hyperparameter summary (Table~\ref{tab2}) given in the main text.

\begin{figure}
  \centering
    \includegraphics[width=0.9\textwidth]{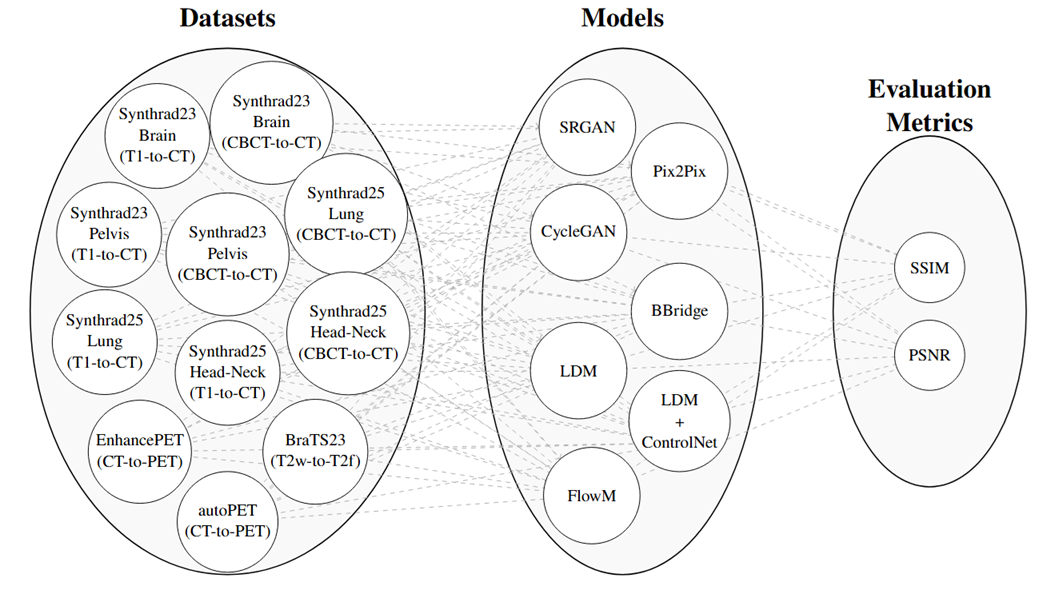}
    \caption{\textbf{The proposed benchmark experiments.} Each of the 11 dataset configurations (left) is evaluated against all 7 generative models (centre) using 2 evaluation metrics (right), yielding 77 experimental combinations in total.}\label{fig2}
\end{figure}

\subsection{Pix2Pix}
Pix2Pix~\cite{isola2017} is a conditional GAN composed of a U-Net generator
and a PatchGAN discriminator that evaluates target realism conditioned
on the source volume. The choice of $\beta_1=0.5$ follows the original
Pix2Pix recommendation, which found lower momentum to stabilize GAN
training. Mixed-precision training with automatic gradient scaling is used
throughout. To stabilize training, generated volumes are not used
immediately to update the discriminator; instead, they are stored in a pool
of 50 previously generated volumes, from which one is randomly selected at
each iteration, exposing the discriminator to a history of generated
samples rather than only the most recent one.

\subsection{CycleGAN}
CycleGAN~\cite{zhu2017} extends Pix2Pix into a bidirectional
image-translation framework, training two ResNet-based generators and two
PatchGAN discriminators simultaneously. The implementation adopted here
extends the original 2D framework to three-dimensional volumetric data,
operating directly on 3D patches throughout all components of the
architecture. All four networks are optimized independently; the learning
rate schedule and mixed-precision settings are identical to Pix2Pix.

\subsection{SRGAN}
SRGAN~\cite{ha2025} is a multi-resolution 3D GAN designed specifically for
volumetric medical image synthesis. The generator is a three-dimensional,
multi-resolution Dense-Attention U-Net (3D-mDAUNet), organized as an
encoder-decoder architecture with skip connections at each resolution
level. The discriminator is a multi-resolution U-Net that produces
voxel-wise real/fake predictions at four spatial scales, providing
spatially dense adversarial feedback at each resolution level. Separate
gradient scalers are used for the generator and discriminator to ensure
numerical stability in their independent update steps.

\subsection{Shared Variational Autoencoder}
\label{sec:supp:vae}
All diffusion and flow-based models operate in a compressed latent space
rather than directly in the voxel domain, made possible by a shared
three-dimensional VAE~\cite{kingma2014}, pre-trained independently on each
target modality and kept frozen during all subsequent model training. The
VAE follows an encoder-decoder architecture based on the AutoencoderKL
framework from the MONAI library~\cite{monai}. Both the encoder and decoder
are organized into three hierarchical stages, with channel widths of 32,
64, and 128 respectively. For a patch of $96\times96\times96$ voxels, the
encoder produces a latent volume of $24\times24\times24$ voxels,
corresponding to a $\times4$ spatial compression factor.

\subsection{LDM-Palette}
\label{sec:supp:ldm}
LDM-Palette~\cite{saharia2022} performs the diffusion process entirely in
the compressed latent space of the frozen VAE (Section~\ref{sec:supp:vae}).
Both the source and target volumes are independently encoded in latent representations. The source latent is used as spatial
conditioning by concatenating it with the noisy target latent channel-wise
before it is passed to the denoising network; since each latent has 3
channels, the concatenated input has 6 channels in total. This spatial
concatenation conditioning avoids the need for cross-attention, making the
conditioning computationally efficient.

Noise is added to the target latent $z_0$ following the DDPM formulation:
\begin{equation}
  z_t = \sqrt{\bar{\alpha}_t}\,z_0
      + \sqrt{1 - \bar{\alpha}_t}\,\varepsilon,
  \quad \varepsilon \sim \mathcal{N}(0, I)
  \label{eq:ddpm_forward_supp}
\end{equation}
where $\bar{\alpha}_t$ is the cumulative product of the noise schedule
coefficients at timestep $t$. A scaled-linear beta schedule is used, with
$\beta_\mathrm{start}=0.0015$, $\beta_\mathrm{end}=0.0205$, and $T=1000$
total timesteps.

\subsection{Brownian Bridge}
\label{sec:supp:bbdm}
The Brownian Bridge diffusion model~\cite{li2023} reformulates the
diffusion process as a stochastic bridge that transitions directly between
the source and target latent representations, without requiring a Gaussian
noise prior at either endpoint. Both the source and target volumes are
first encoded into their respective latent representations using the
frozen VAE encoder (Section~\ref{sec:supp:vae}). The forward process
defines a noisy interpolated trajectory between the target latent $z_0$ and
the source latent $z_y$ at each discrete timestep $t$:
\begin{equation}
  z_t = (1 - m_t)\,z_0 + m_t\,z_y + \sigma_t\,\varepsilon,
  \quad \varepsilon \sim \mathcal{N}(0, I)
  \label{eq:bbdm_forward_supp}
\end{equation}
where $m_t$ is a mixing coefficient that increases linearly from $0.001$ to
$0.999$ over $T=1000$ timesteps, so that at $t=0$ the process is close to
the target latent and at $t=T$ it is close to the source latent. The noise
variance at each step is $\sigma_t^2 = 2(m_t - m_t^2)$, which peaks at the
midpoint of the trajectory and vanishes at both endpoints, ensuring a
smooth and bounded stochastic interpolation. An exponential moving average
of the model weights is maintained with a decay of 0.995, starting from
step 1000 and updated every 8 steps.

\subsection{LDM-Palette + ControlNet}
ControlNet~\cite{zhang2023} augments a pre-trained diffusion model with a
trainable auxiliary network that injects spatially structured conditioning
information via residual connections, without modifying the backbone
weights. Here, ControlNet is applied on top of the frozen LDM-Palette
backbone (Section~\ref{sec:supp:ldm}), extending the implementation
proposed in~\cite{moschetto2025} from 2D slice-wise processing to fully
volumetric 3D inference in the latent space of the shared VAE, with only
the ControlNet parameters updated during training. As a training
diagnostic, the magnitude of the ControlNet residuals is monitored every 50
epochs to verify that the module contributes meaningfully to the denoising
process without destabilizing training. The diffusion schedule and
inference procedure are identical to LDM-Palette, using DDIM sampling over
50 steps starting from Gaussian noise in the latent space.

\subsection{Flow Matching}
Flow Matching~\cite{lipman2022} offers a deterministic alternative to
stochastic diffusion by learning a time-varying velocity field that
transports samples along straight-path trajectories directly from source to
target. Unlike LDM-Palette and ControlNet, which predict the noise added to
the target latent at each timestep and require a stochastic reverse process
guided by a noise scheduler, Flow Matching predicts the velocity, i.e., the
direction and rate of change needed to move from the source latent toward
the target latent in a single continuous flow. Inference is therefore fully
deterministic and does not require a noise scheduler, as integration starts
directly from the source latent rather than from Gaussian noise. The
implementation adopted here extends~\cite{moschetto2025} from 2D
slice-wise processing to fully volumetric 3D inference in the latent space
of the shared VAE. Unlike the other diffusion-based models, Euler
integration is implemented directly in the trainer rather than relying on
an external inference scheduler, keeping the sampling procedure
self-contained and independent of the diffusion library infrastructure.

\section{Turing platform}\label{app:turing_platform}

\begin{figure}
  \centering
    \includegraphics[width=\textwidth]{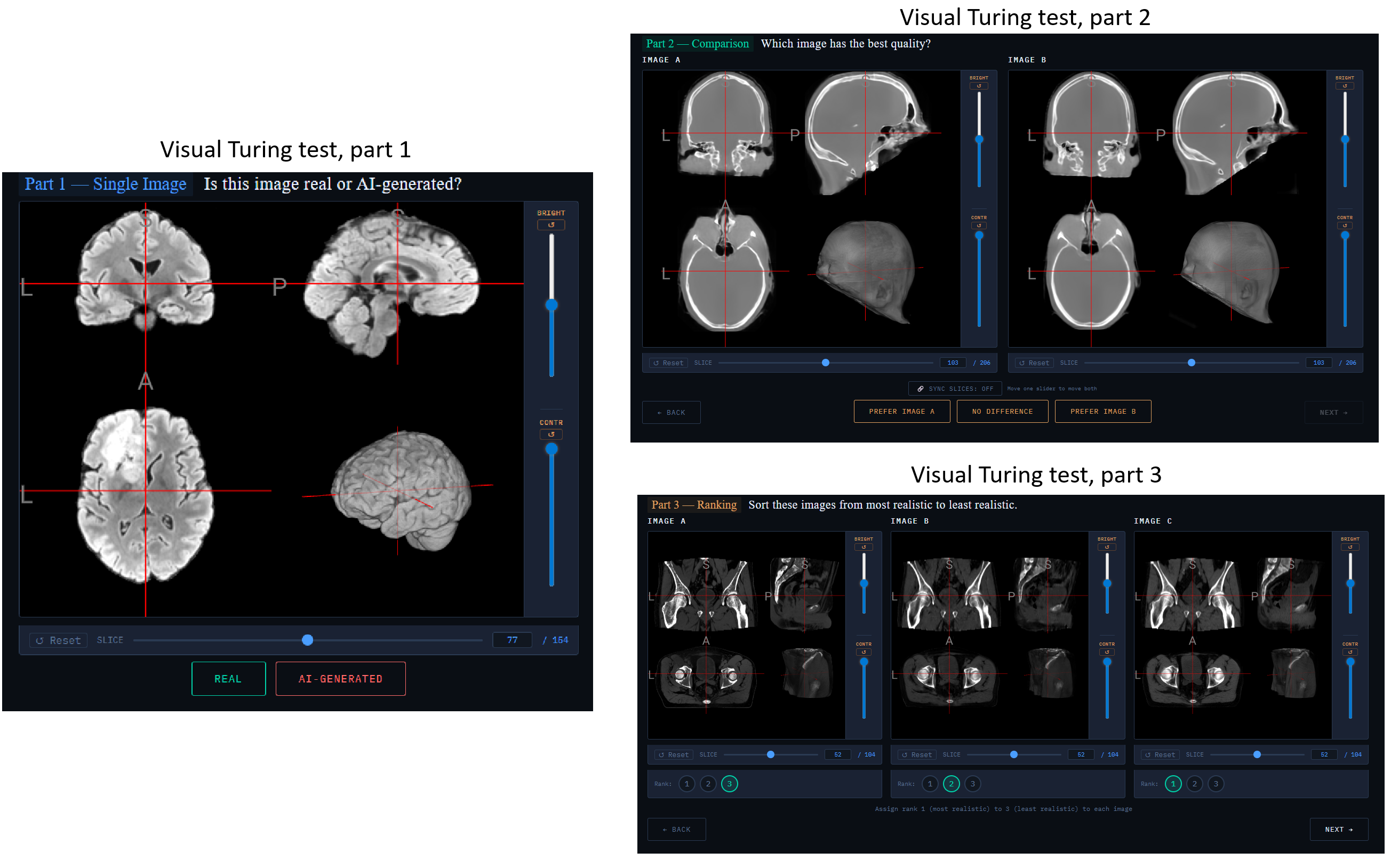}
    \caption{\textbf{The Visual Turing test platform.} Each volume is displayed through a multi-planar viewer rendered by a grid layout
providing three orthogonal anatomical planes (axial, sagittal, and coronal) alongside a 3D surface reconstruction. In Part 1, a single volume is displayed and participants are asked to classify the image as either \textit{Real} or \textit{AI-generated} using two mutually exclusive buttons positioned below the viewer.
    In Part 2, two volumes from the same patient are displayed side by
side, each with independent brightness, contrast, and slice controls. Participants select \textit{Prefer Image A}, \textit{Prefer Image B}, or \textit{No Difference}. An optional toggle links the slice positions of the two viewers, enabling direct
anatomical comparison at the same relative depth across both images simultaneously.
In Part 3, three volumes from the same patient are displayed simultaneously in a three-column grid, each with independent viewer controls.
Participants assign a rank from 1 (most realistic) to 3 (least realistic) to each image using circular rank buttons below each viewer. 
} \label{figC}
\end{figure}

The Visual Turing Test has been administered through a custom web platform specifically designed for the interactive review of 3D medical volumes (Figure \ref{figC}). 
The platform is implemented as a client-side JavaScript application served through JATOS, enabling deployment on any institutional server without external dependencies. 
Volumes are rendered using NiiVue v0.50.0 \cite{niivue}, which provides WebGL-accelerated multi-planar reformatting and 3D surface reconstruction directly in the browser. 
The planar views are spatially linked, allowing simultaneous navigation across all anatomical planes.
Supported question types include:
(i) single-volume binary classification (real vs. synthetic);
(ii) side-by-side pairwise preference with optional synchronized slice navigation; and (iii) multi-volume ordinal ranking with drag-and-drop or button-based rank assignment. 
Control questions (identical volume
pairs or triplets) have been inserted at arbitrary positions to monitor response consistency.  
All responses have been stored server-side in CSV format,
anonymized by participant ID, and include timestamps for response-time analysis. 

Upon accessing their personalized link, participants are directed to a landing page where they report their years of clinical experience (${<}5$~years / 5–10~years / ${>}10$~years) and primary imaging modality (CT, MRI, PET, or all modalities).
Following the completion of the test, some participants provided feedback highlighting aspects of the study design that may have influenced their responses, and that should be considered when designing future reader studies for medical image synthesis.
Specifically, participants could not adjust the window width and level of the displayed volumes; all images were presented in a grayscale color-map; no maximum intensity projection (MIP) views were provided; and all images were shown in isolation, whereas in clinical practice PET is interpreted as a hybrid examination in combination with CT or MRI. 
These factors have been taken into account when interpreting the Turing test results, and future  studies would benefit from incorporating adjustable windowing, clinically standard color maps, MIP reconstructions, and fused multi-modal displays to better align with the diagnostic reading environment.

\section{Additional results}\label{app:additional_results}

\begin{figure}[h]
  \centering
    \includegraphics[width=\textwidth]{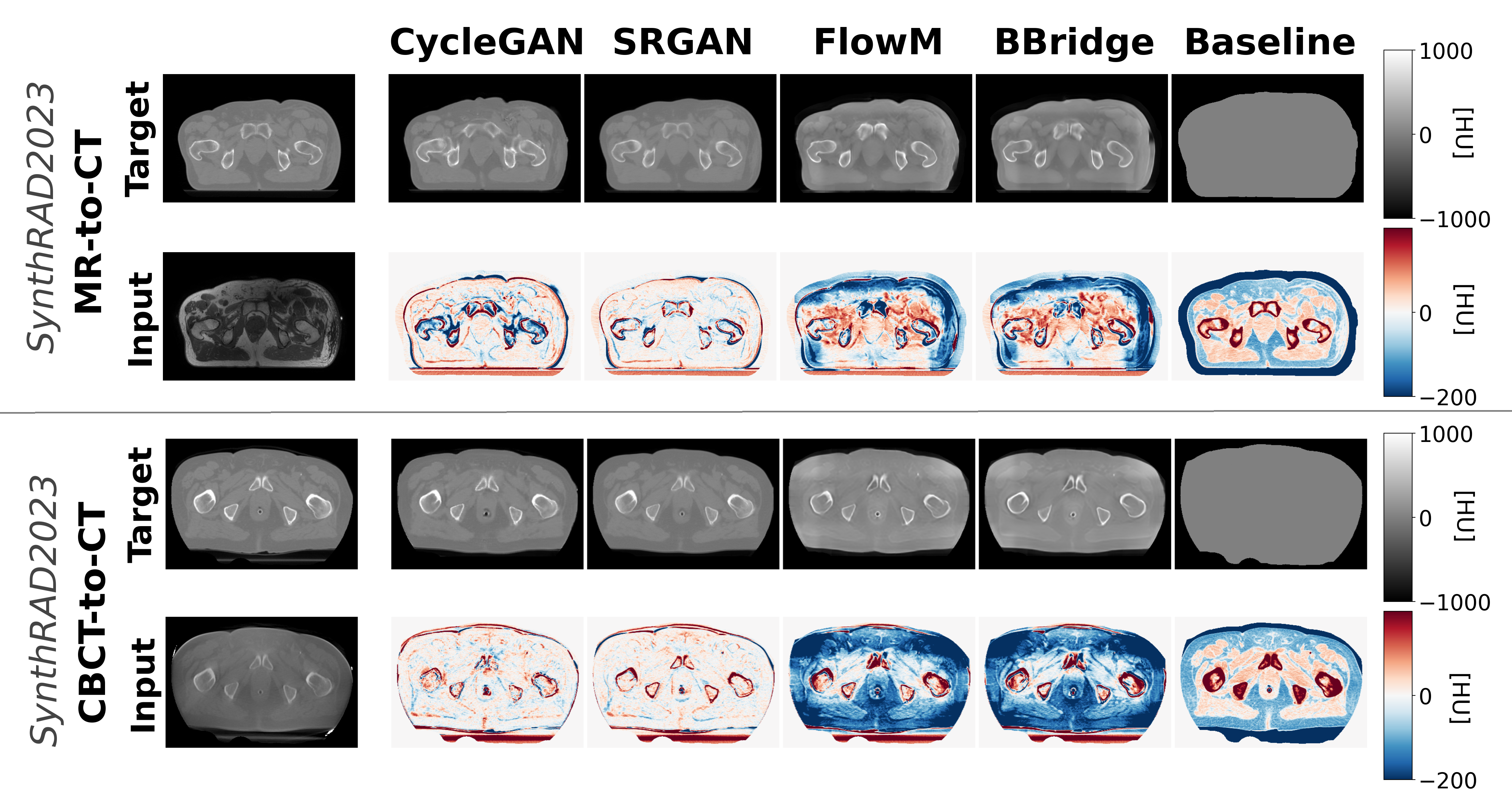}
    \caption{Visual comparison across \textbf{pelvis} I2I translation tasks, for the two best-performing GANs (SRGAN and CycleGAN) and latent generative models (BBridge and FlowM). 
    For each task, we display the target and input images (first column, first and second row respectively); the corresponding model predictions (first row); and 
    the associated error maps with respect to the reference target (second row), computed as the pixel-wise difference between the synthesized output and the ground truth.
    The last column includes a task-specific baseline, representing a prediction with no learned structure. The baseline image is a uniform water image (0 HU) for CT tasks and a mean intensity image for the other modalities.}\label{fig:pelvis}
\end{figure}

\begin{figure}[h]
  \centering
    \includegraphics[width=\textwidth]{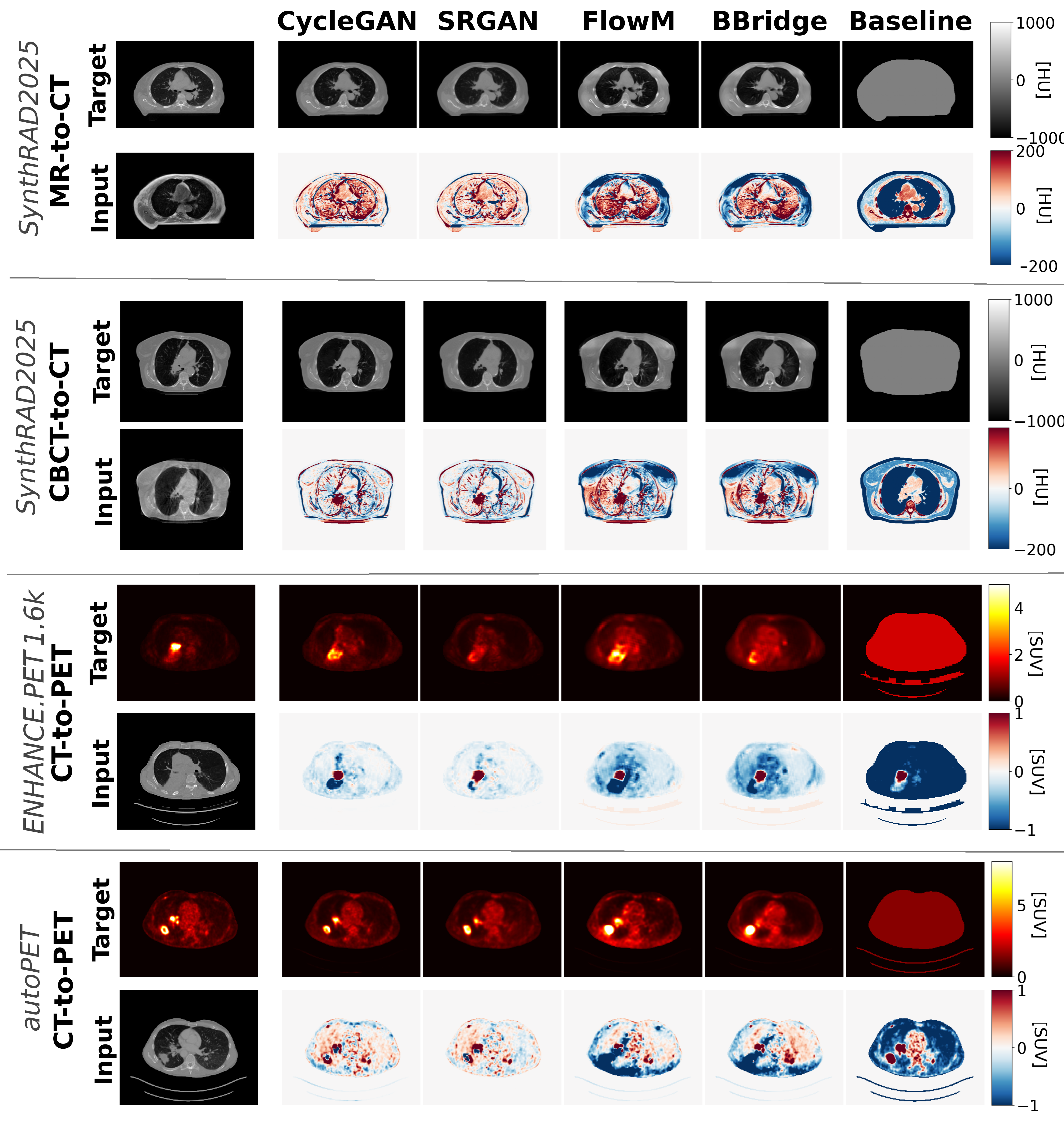}
    \caption{The same as Figure~\ref{fig:pelvis}, but for \textbf{lung}.}\label{fig:lung}
\end{figure}

\begin{figure}[h]
  \centering
    \includegraphics[width=\textwidth]{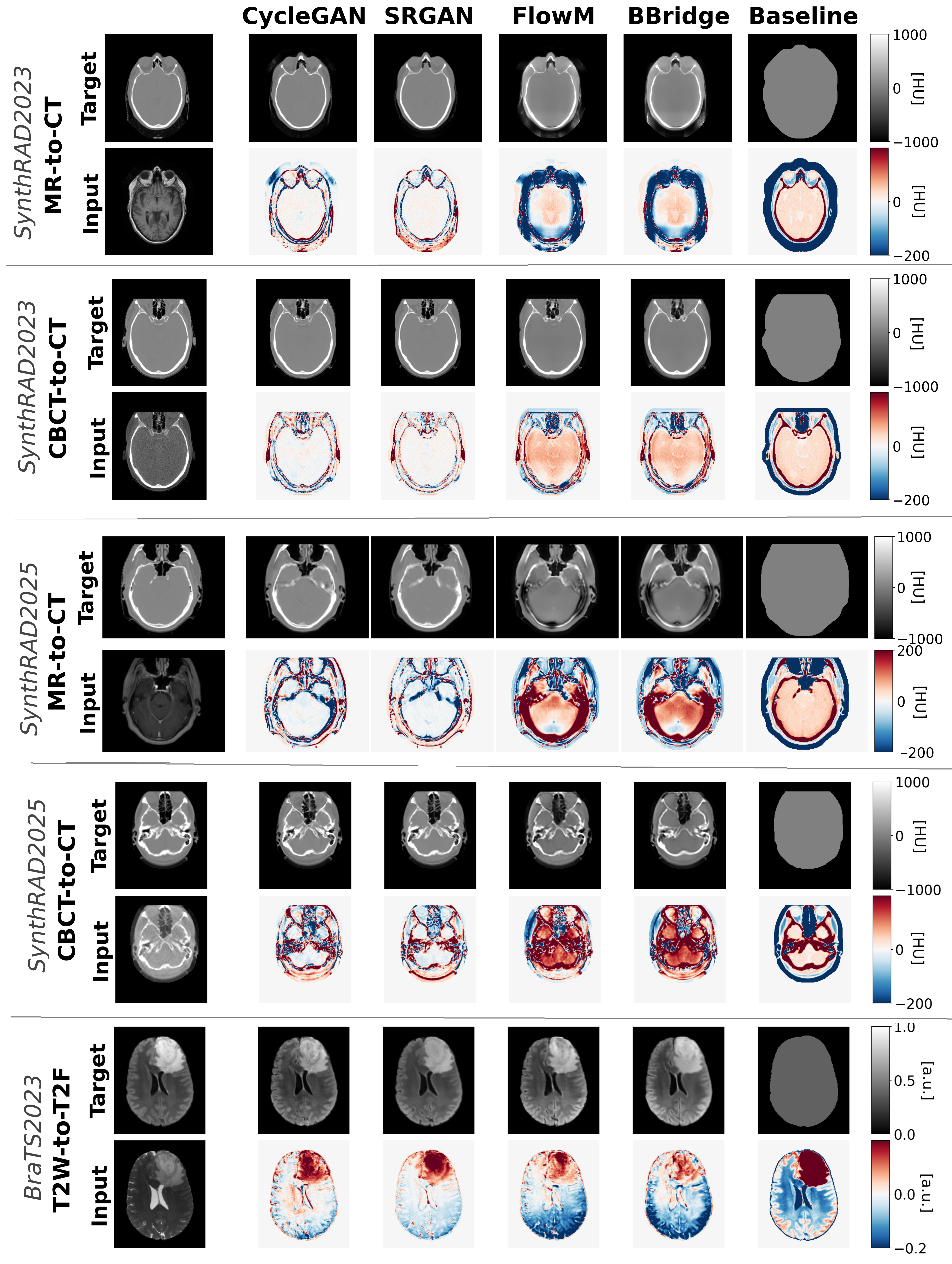}
    \caption{The same as Figure~\ref{fig:pelvis}, but for \textbf{brain} and \textbf{head/neck}.}\label{fig:brain}
\end{figure}

Figures~\ref{fig:pelvis},~\ref{fig:lung}, and~\ref{fig:brain} compare the images generated from the two best-performing GAN-based (SRGAN and CycleGAN) and latent generative models (BBridge and FlowM), across all tasks and datasets. 
For each task, the input and target images are displayed alongside the model predictions and the corresponding error maps, computed as the pixel-wise difference between the synthesized output and the ground truth. 
A task-specific baseline is also included, representing a prediction with no learned structure: a uniform water image (0 HU) for CT tasks and a mean intensity image for the other modalities.
Across all tasks, GAN-based methods exhibit lower and more spatially confined errors compared to latent models.
Models' performance is influenced by the heterogeneous tissue composition of the anatomical regions. 
In anatomically complex regions, such as the nasal cavities, sinuses, and dental areas in head/neck tasks (Figure~\ref{fig:brain}), or the bone–soft tissue interfaces in the pelvis (Figure~\ref{fig:pelvis}), errors are larger than in more homogeneous regions like the brain parenchyma, where residuals remain low and uniform.
The largest errors occur in correspondence with lesion regions, whose irregular morphology and appearance make them particularly difficult to reproduce, and at image boundaries, when transitioning to the background. 
For instance, in the T2w-to-T2f brain task (BraTS2023, Figure~\ref{fig:brain}), all models exhibit elevated residuals at the tumor site compared to the surrounding healthy tissue.
This difficulty is further amplified in inter-modality tasks such as CT-to-PET, where inferring functional metabolic activity from anatomical information requires bridging a full domain shift.

\section{Performance Evaluation}\label{app:performance}

Inference time was measured on a single NVIDIA GH200 GPU (96GB) (PyTorch 2.5.1, CUDA 12.4) on 10 test volumes per task. 
Each measurement covers the full inference pipeline of a model on one volume (foreground cropping, sliding-window inference over $96\times 96 \times 96$ patches and, for latent models, iterative sampling with 50 steps, 200 for BBridge), and excludes data loading and file saving. 

\begin{table}
\caption{Inference time per test volume (seconds; mean with standard deviation as subscript over 10 test volumes per task). The last rows report the median over tasks, the peak GPU memory and the number of parameters used at inference.}\label{tbl_inference_time}
\fontsize{8}{9.5}\selectfont
\setlength{\tabcolsep}{2pt}
\begin{tabular*}{\tblwidth}{@{\extracolsep{\fill}}lllccccccc}
\toprule
\textbf{District} & \textbf{Task} & \textbf{Dataset} & \textbf{SRGAN} & \textbf{CycleGAN} & \textbf{Pix2Pix} & \textbf{BBridge} & \textbf{LDM} & \textbf{LDM+CN} & \textbf{FlowM} \\
\midrule
Brain & T2w-to-T2f & BraTS23 & $1.00_{0.26}$ & $0.23_{0.06}$ & $0.03_{0.01}$ & $51.4_{13.4}$ & $11.2_{2.85}$ & $15.2_{3.95}$ & $12.0_{3.01}$ \\
Head-Neck & T1w-to-CT & Synthrad23 & $2.83_{0.75}$ & $0.67_{0.18}$ & $0.12_{0.04}$ & $146.2_{38.8}$ & $30.7_{7.86}$ & $44.2_{11.7}$ & $32.1_{8.48}$ \\
Head-Neck & T1w-to-CT & Synthrad25 & $5.59_{0.28}$ & $1.32_{0.07}$ & $0.20_{0.02}$ & $294.2_{14.9}$ & $57.5_{4.92}$ & $86.5_{4.35}$ & $66.1_{5.53}$ \\
Head-Neck & CBCT-to-CT & Synthrad23 & $4.20_{1.43}$ & $1.01_{0.35}$ & $0.14_{0.05}$ & $220.6_{75.3}$ & $53.5_{18.3}$ & $65.1_{22.2}$ & $47.0_{16.0}$ \\
Head-Neck & CBCT-to-CT & Synthrad25 & $1.02_{0.16}$ & $0.24_{0.03}$ & $0.05_{0.03}$ & $54.6_{8.13}$ & $12.0_{1.75}$ & $15.7_{2.35}$ & $11.3_{1.66}$ \\
Lung & CT-to-PET & autoPET & $0.83_{0.39}$ & $0.20_{0.09}$ & $0.04_{0.02}$ & $42.9_{20.3}$ & $9.64_{4.56}$ & $12.7_{5.99}$ & $9.44_{4.45}$ \\
Lung & CT-to-PET & EnhancePET & $0.14_{0.04}$ & $0.04_{0.01}$ & $0.01_{0.01}$ & $7.04_{2.47}$ & $1.71_{0.55}$ & $2.13_{0.71}$ & $1.55_{0.50}$ \\
Lung & T1w-to-CT & Synthrad25 & $2.56_{0.67}$ & $0.61_{0.16}$ & $0.09_{0.02}$ & $132.5_{35.0}$ & $22.9_{3.62}$ & $39.5_{10.5}$ & $29.0_{7.66}$ \\
Lung & CBCT-to-CT & Synthrad25 & $0.97_{0.33}$ & $0.24_{0.07}$ & $0.04_{0.01}$ & $50.2_{17.3}$ & $11.4_{3.87}$ & $15.1_{5.23}$ & $12.1_{4.03}$ \\
Pelvis & T1w-to-CT & Synthrad23 & $7.14_{1.98}$ & $1.70_{0.47}$ & $0.24_{0.07}$ & $366.1_{101.5}$ & $71.8_{21.6}$ & $109.7_{30.3}$ & $80.1_{22.2}$ \\
Pelvis & CBCT-to-CT & Synthrad23 & $3.56_{1.34}$ & $0.84_{0.32}$ & $0.13_{0.07}$ & $184.0_{69.0}$ & $41.8_{15.8}$ & $55.0_{20.6}$ & $40.5_{15.2}$ \\
\midrule
\multicolumn{3}{l}{Median over tasks (s)} & 2.56 & 0.61 & 0.09 & 132.5 & 22.9 & 39.5 & 29.0 \\
\multicolumn{3}{l}{Peak GPU memory (GiB)} & 2.6 & 1.5 & 0.9 & 7.9 & 4.8 & 16.8 & 16.1 \\
\multicolumn{3}{l}{Parameters (M)} & 6 & 9 & 67 & 671 & 475 & 655 & 475 \\
\bottomrule
\end{tabular*}
\end{table}

The two families differ in computational cost (Table~\ref{tbl_inference_time}). 
GAN-based models translate a volume under a few seconds (median over tasks 0.09~s for Pix2Pix, 0.61~s for CycleGAN and 2.56~s for SRGAN), with a peak GPU memory of at most 2.6~GiB. 
Latent generative models are one to two orders of magnitude slower, requiring a median of 22.9~s (LDM), 29.0~s (FlowM), 39.5~s (LDM+ControlNet) and 132.5~s (BBridge) per volume, and up to 16.8~GiB of GPU memory. 
This gap reflects both the iterative sampling required for each patch during sliding-window inference (50 steps, or 200 for BBridge) and the larger networks (475--671 million parameters, compared with 6--67 million for GANs).
Inference time scales with the volume size: for BBridge it ranges from 7.0~s on the cropped EnhancePET volumes to 366~s on Synthrad23 T1w-to-CT (pelvis). 
Therefore, in addition to achieving the highest image fidelity, SRGAN provides 9--52-fold faster inference than the latent models, offering a practical advantage in time-sensitive workflows such as online adaptive radiotherapy.

\end{document}